\documentclass[sigconf,10pt]{acmart}

\usepackage{times}
\usepackage{soul}
\usepackage{url}
\usepackage[utf8]{inputenc}
\usepackage{caption}
\usepackage{tikz}
\usepackage{amsmath,amsfonts, bm}
\usepackage{algorithm}

\usepackage{algorithmicx}
\usepackage{algpseudocode}

\usepackage{graphicx}
\usepackage{subcaption}		
\usepackage{textcomp}
\usepackage{booktabs} 
\usepackage{siunitx}
\usepackage{xspace,url}

\usepackage{color}
\usepackage{colortbl}
\usepackage{multirow}
\usepackage{multicol}
\usepackage{makecell}
\usepackage{tabularx}
\usepackage{listings}
\usepackage{graphicx}
\usepackage{caption}
\usepackage{enumitem}
\usepackage{adjustbox}

\usepackage{filecontents}
\usepackage[colorinlistoftodos]{todonotes}

\acmYear{2023}\copyrightyear{2023}
\setcopyright{rightsretained}
\acmConference[Preprint]{}{}{2023}

\definecolor{airforceblue}{rgb}{0.36, 0.54, 0.66}
\newcommand{\name}{NN-Factory\xspace} 

\newcommand{\ie}{\textit{i}.\textit{e}.~}
\newcommand{\eg}{\textit{e}.\textit{g}.~}

\title{Generative Model for Models: Rapid DNN Customization for Diverse Tasks and Resource Constraints}

\author{Wenxing Xu}
\authornote{Work was done while the authors were interning at Institute for AI Industry Research (AIR), Tsinghua University.}
\affiliation{Beijing University of Posts and Telecommunications}

\author{Yuanchun Li}
\authornote{Corresponding author.}
\affiliation{Institute for AI Industry Research (AIR), Tsinghua University}

\author{Jiacheng Liu}
\authornotemark[1]
\affiliation{Beijing Institute of Technology}

\author{Yi Sun}
\affiliation{Institute for AI Industry Research (AIR), Tsinghua University}

\author{Zhengyang Cao}
\authornotemark[1]
\affiliation{University of Electronic Science and Technology of China}

\author{Yixuan Li}
\authornotemark[1]
\affiliation{Beijing University of Posts and Telecommunications}

\author{Hao Wen}
\affiliation{Institute for AI Industry Research (AIR), Tsinghua University}

\author{Yunxin Liu}
\affiliation{Institute for AI Industry Research (AIR), Tsinghua University}

\renewcommand{\shortauthors}{Xu and Li et al.}

\setcopyright{acmcopyright}
\renewcommand\footnotetextcopyrightpermission[1]{}

\begin{document}




\begin{abstract}
Unlike cloud-based deep learning models that are often large and uniform, edge-deployed models usually demand customization for domain-specific tasks and resource-limited environments. Such customization processes can be costly and time-consuming due to the diversity of edge scenarios and the training load for each scenario. Although various approaches have been proposed for rapid resource-oriented customization and task-oriented customization respectively, achieving both of them at the same time is challenging. Drawing inspiration from the generative AI and the modular composability of neural networks, we introduce \name, an one-for-all framework to generate customized lightweight models for diverse edge scenarios. The key idea is to use a generative model to directly produce the customized models, instead of training them. The main components of \name include a modular supernet with pretrained modules that can be conditionally activated to accomplish different tasks and a generative module assembler that manipulate the modules according to task and sparsity requirements. Given an edge scenario, \name can efficiently customize a compact model specialized in the edge task while satisfying the edge resource constraints by searching for the optimal strategy to assemble the modules. Based on experiments on image classification and object detection tasks with different edge devices, \name is able to generate high-quality task- and resource-specific models within few seconds, faster than conventional model customization approaches by orders of magnitude.
\end{abstract}

\maketitle

\section{Introduction}

Deep learning (DL) has become a game-changing artificial intelligence (AI) technology in recent years.
It has achieved remarkable performance in various domains including computer vision, natural language understanding, computer gaming, and computational biology.
Meanwhile, deep learning has enabled and enhanced many intelligent applications at the edge, such as driving assistance \cite{VIeye,mobiad}, face authentication \cite{face_infocom}, video surveillance \cite{loki,jiang2021remix}, speech recognition \cite{google_speech,nipsspeech}, etc.
Due to latency and privacy considerations, it is becoming an increasingly common practice to deploy the models to edge devices \cite{liu2020pmc,mistify,convrelu}, so that the models can be directly invoked without transmitting data to the server.

To adopt DL models in different edge scenarios, developers usually need to customize the models, which includes the following two main processes.
\begin{enumerate}
\item \textbf{Task-oriented Customization}\footnote{In this paper, we focus on the simplified case when the tasks across different edge scenarios share a combinatorial space. For example, each task is to classify/detect objects with a combination of several known properties (\eg color, type, category, element, etc.).}:
Customizing the models for domain-specific tasks, such as detecting certain types of cars, tracking people with certain costume, or identifying items with certain failures.
\item \textbf{Resource-oriented Customization}: Customizing the models for edge devices, in order to fulfill certain resource constraints on the target edge device, including the memory budget and the latency requirement.
\end{enumerate}
The current practice is to deal with each customization problems separately. For example, developers need to generate compact models through extensive model architecture designing/compression to meet the resource constraints, and then train/fine-tune the model on the domain-specific tasks to improve accuracy.
Such a process may involve expensive data collection, time-consuming architecture search, and unstable model training.

Recent advances of deep learning have demonstrated excellent zero-shot adaptation abilities of large pretrained models, including training-free neural architecture adaptation and task adaptation.
Specifically, one-shot neural architecture search (NAS) \cite{ofa,proxylessnas} and model scaling \cite{legodnn,nestdnn,wen2023adaptivenet,harvnet} techniques have demonstrated the possibility of training a supernet and customizing it for diverse edge environments.
Pretrained large models like ChatGPT \cite{chatgpt}, OFA \cite{oneforall}, and Segment Anything \cite{PaLM} allow users to customize the task of the model by simply providing a task prompt as the model input.
Such training-free adaptation abilities are desirable in edge AI scenarios due to the possibility to support a wide range of downstream tasks and diverse hardware constraints with significantly reduced development efforts.

However, it is difficult to combine training-free neural architecture customization and training-free task customization together.
On one hand, using one-shot NAS or model scaling to generate models for different tasks is not feasible since the supernet is designed for a single task. Extending them to multiple tasks would significantly enlarge the model search space, making the supernet training and subnet search extremely challenging.
On the other hand, the remarkable task adaptation abilities of pretrained models come at the cost of huge computational load. Recent studies have shown that the zero-shot generalization ability of pretrained models emerges and strengthens with increasing model size \cite{kaplan2020scaling}. Thus, it is difficult or even impossible to directly use the task-generalizable large model due to the diverse and limited edge computational resources.

To achieve both task-oriented and resource-oriented efficient model customization, we introduce \name, a new paradigm for generating customized models for diverse edge scenarios.
Our key idea is to \emph{view the problem of edge-specific model customization as a generative problem - rather than existing generative models that are designed for generating media content \cite{chatgpt,creswell2018generative,croitoru2023diffusion}, we use generative AI to produce customized models based on edge scenario specifications}.
Specifically, given the desired task, the device type, and the resource requirements in an edge scenario, \name rapidly produces a compact model that matches the task and requirements by simply querying the generative model.
Such a new paradigm of model customization has the potential to fundamentally avoid the cumbersome compression and training processes in conventional edge model customization approaches.


\begin{figure}
    \centering
    \includegraphics[width=0.47\textwidth]{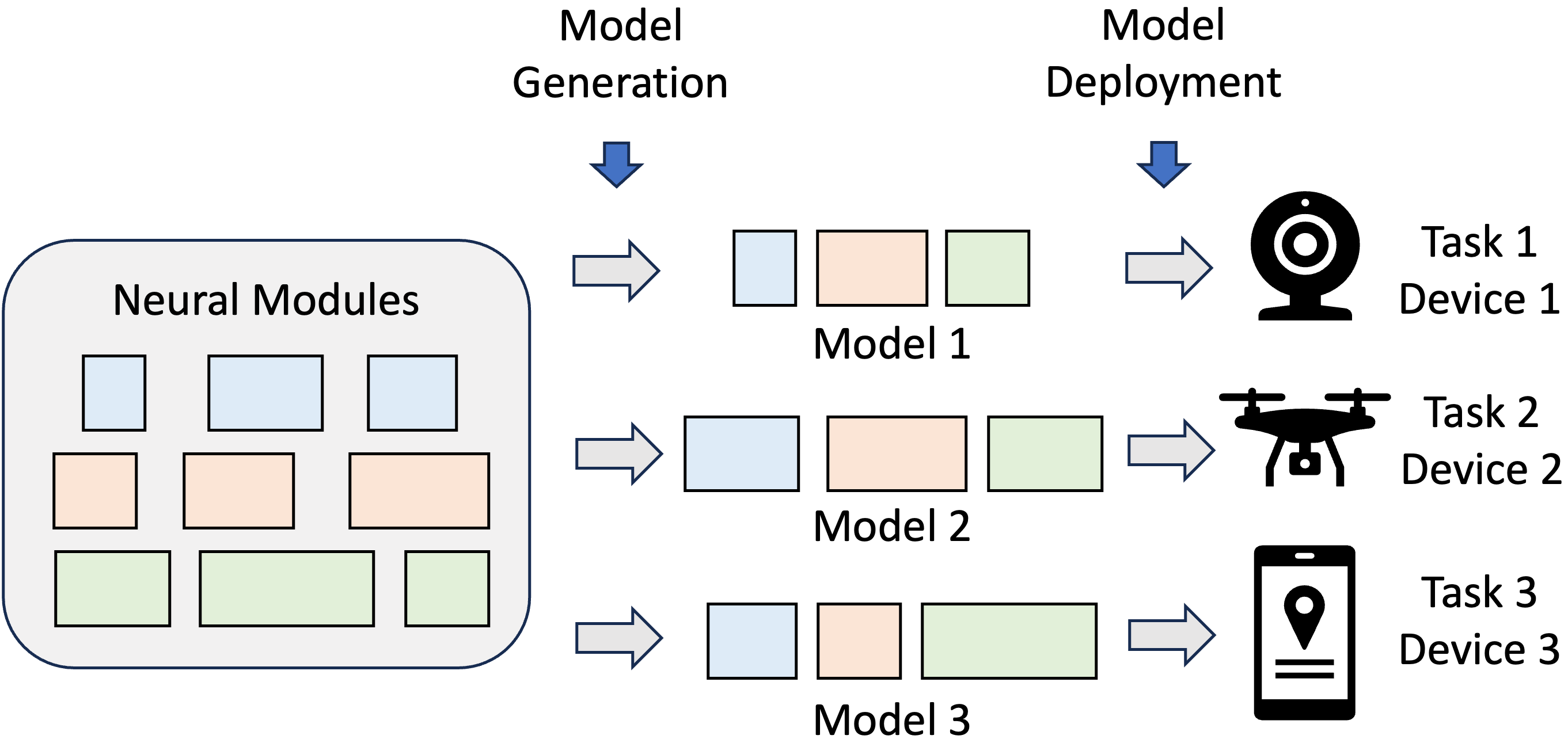}
    \caption{The key idea of \name: Generating models for diverse edge scenarios by assembling modules.}
    \label{fig:illustration}
\end{figure}

However, directly generating the parameters of a model is difficult due to the enormous number of model parameters. The key insight of \name to address this problem is the composability of neural network modules \cite{andreas2016neural,Neuro-symbolic}, \ie different functionalities and sizes of neural networks can be achieved by certain combinations of neural network modules, as illustrated in Figure~\ref{fig:illustration}. Such modular composability has been studied by existing approaches (\eg NestDNN \cite{nestdnn}, LegoDNN \cite{legodnn}, AdaptiveNet \cite{wen2023adaptivenet}, etc.) for edge-side model scaling, but none of them is able to support task-oriented customization.

Based on this insight, \name generates a customized model for each edge scenario by directly generating the configurations to assemble pretrained neural modules. The key components of \name include a modular supernet that carries the neural modules, 
a task- and sparsity-aware module assembler that generates candidate module assembling configurations (each configuration is a set of gate vectors that can map to a candidate model), and a lightweight architecture searcher that rapidly finds the optimal configuration to assemble the model for the target edge scenario.

Specifically, the modular supernet is extended from a common backbone network, such as Convolutional Neural Network (CNN) or Transformer, in which each basic block is decomposed to multiple conditionally activated modules.
How the modules are activated is controlled by several gate vectors, which are produced by a generative model (\ie the module assembler). By activating different sets of modules, the basic blocks in the backbone network can be reconfigured to have different functionalities and sizes.
The module assembler decides how to activate the modules in the backbone network based on a task description and a sparsity requirement (\ie a limit of module activation ratio). 
The modules and assembler are jointly trained to coherently work together.

Based on the module-assembler design, we are able to reduce the large search space of module combinations to a small search space of generation prompts.
To find the customized model for an edge scenario (defined by the task, device, and latency/memory requirements), we only need to find the optimal model generation request, which is done by the lightweight architecture search module under the guidance of a device-specific model performance evaluator.
Due to the significantly reduced search space, finding the optimal model only takes few short iterations.

To evaluate \name, we conduct experiments with three devices on two types of tasks. The results have demonstrated that our approach can generate tailored models for a given edge scenario in up to 6 seconds, 1000$\times$ faster than the conventional training-based customization approach. The generated models can achieve high accuracy that is comparable with edge retrained/fine-tuned models on both known and unseen tasks. 

Our work makes the following technical contributions:
\begin{enumerate}
    \item We propose a novel generative AI-based edge-specific model customization solution. It enables rapid training-free model customization for edge scenarios with diverse tasks and resource constraints.
    \item We introduce a modular design for model generation, which enables rapid model customization by simply querying an assembler model for module activations.
    \item Based on experiments with two types of tasks and various edge devices, our method is able to generate high-quality models for diverse edge scenarios with significantly lower overhead. The system and models will be open-sourced.
\end{enumerate}

\section{Background and Related Work}
\label{section:background}


\subsection{Model Customization for Edge Scenarios}

Deploying deep neural networks (DNNs) at the edge is increasingly popular due to the latency requirements and privacy concerns of deep learning services.
However, directly following the common cloud-based model training and deployment procedures is not satisfactory due to the huge diversity of edge scenarios \cite{wen2023adaptivenet,legodnn,nestdnn}.
Unlike most cloud-based AI models that are designed for generic tasks and hosted in powerful GPU clusters, edge AI scenarios are usually diverse and fragmented.
Each edge scenario may deal with a domain-specific task and a unique set of target deployment environments.

\begin{figure}
    \centering
    \includegraphics[width=0.47\textwidth]{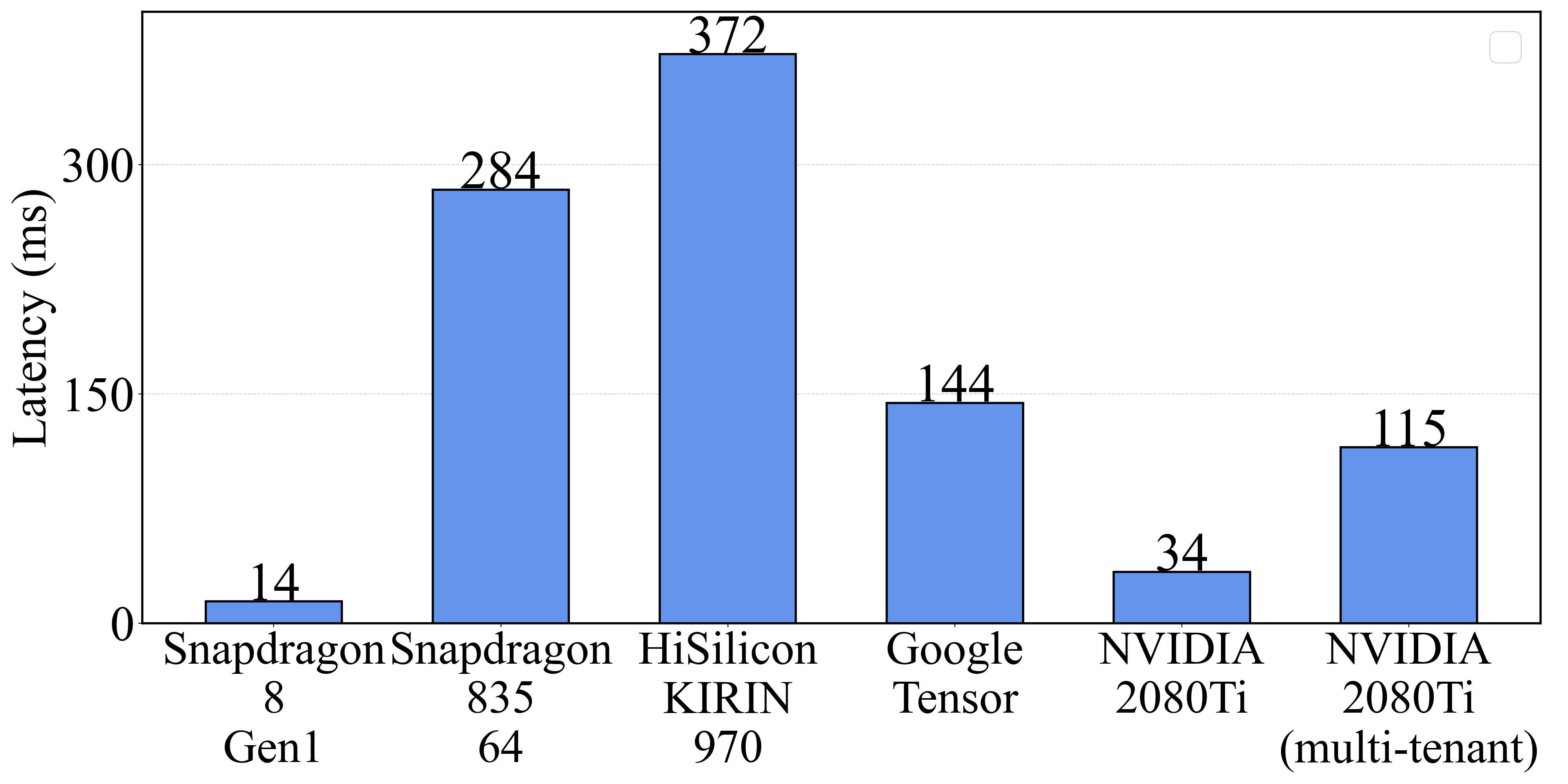}
    \caption{Average latency (ms) of ResNet50 in different deployment environments. In the ``multi-tenant'' setting, we assume there are three background processes running the same model. }
    \label{fig:comparison}
\end{figure}

\begin{table}
	\caption{Typical AI tasks requested by the customers of an edge AI service provider.}
	\centering
	\resizebox{.47\textwidth}{!}
    {
	\begin{tabular}{cl}
		\toprule
		Customer    &  Task  \\
		\midrule
        Construction site & Detect people wearing blue uniform. \\ 
		Restaurant & Classify people wearing white chef cap. \\
        Hospital & Classify people wearing masks. \\
		Shopping mall & Detect smoking people. \\
		Traffic station & Detect cars with green plates. \\
        \bottomrule
	\end{tabular}
	}
	\label{tab:bg:task-diversity}
\end{table}   

According to our industry partners that provide edge AI services for the customizers, their major development efforts are spent on customizing models to handle the diversities of edge scenarios, including hardware diversity, task diversity, data distribution diversity, etc.
Among them, a major problem is the hardware diversity. Customizers usually require to deploy models to different hardware platforms, such as edge servers, desktops, smartphones, and edge AI boxes. Due to the different computational capabilities, the same model may yield significantly different performances on different edge devices and environments, as shown in Figure~\ref{fig:comparison}. Therefore, developers need to customize the model architectures to meet the latency constraints.
Task diversity is another important problem faced by edge AI service providers. Each customizer may have a domain-specific AI task based on the application scenario, as illustrated in Table~\ref{tab:bg:task-diversity}. To enable AI services in such scenarios, customizing the model for the different tasks is necessary.


There are various existing approaches proposed for each type of the above customization goals. However, when considering the two goals jointly for many diverse edge scenarios, the cost becomes huge, since the customization process is non-trivial (as will be explained later) and has to be repeated for each scenario.


\subsection{Training-based Model Customization}

To customize a model for a certain domain-specific task, the common practice is to use transfer learning (TL). 
The predominant approach in transfer learning is is fine-tuning, \ie initializing the model parameters with a pretrained model and continue training the parameters with domain-specific data.
Such transfer learning processes require developers to collect data samples for the task, label them, and train the model with the labeled samples, which are usually labor-intensive, resource-demanding, and time-consuming.

To fit a model into a specific edge device, the typical solutions include compressing an existing model \cite{deep_compression, haq, int8} (\eg pruning, quantization, etc.) or finding a new model architecture that aligns with the capabilities of the target device \cite{mbv2, efficientnetv2}. Since manually designing models for diverse edge environments is cumbersome, the common practice is to use automated model generation techniques. NAS~\cite{mnasnet,single_path_one_shot,fbnet,netadapt,dna} is the most representative and widely-used model generation method, which searches for the optimal network architecture in a well-designed search space.
Most NAS methods require training the architectures during searching \cite{nas-rl, evolutionarysearch, mnasnet, darts-gradient-based}, which is exceedingly time-consuming (10,000+ GPU hours) when generating models for a large number of devices.

Both of the above processes pose challenges for developers due to the diversity of edge scenarios. Training or fine-tuning the model for one edge scenario typically requires thousands of labeled training samples and several hours on high-performance GPU machines. Finding the optimal model architecture or compression strategy by trial and error also requires much development effort.

\subsection{Training-free Model Customization}

The recent advances of deep learning have shown the feasibility to use a unified pretrained model to support various downstream tasks without further training. 
Specifically, one can train a foundation model on a large dataset with various predefined tasks and directly use the model for different downstream tasks with simple prompts.
According to empirical analyses \cite{kaplan2020scaling}, improving the capacity of pretrained model can lead to better performance on downstream tasks. 

Meanwhile, both the AI community and the mobile computing community have attempted various ways to reduce the cost of generating lightweight models for edge devices. One-shot NAS \cite{proxylessnas, ofa, foxnas, sgnas_oneshot} is proposed to greatly reduce the training cost by allowing the candidate networks to share a common over-parameterized supernet. The best models for the target device can be found by directly searching a subnet in the supernet.
Mobile computing researchers have also proposed to scale models dynamically to provide a wide range of resource-accuracy trade-offs. 
Most of them apply structured pruning or model architecture adaptation to generate descendent models \cite{nestdnn, slim, us_slim, legodnn, network_slimming} with different levels of computational cost.

Training-free model adaptation can also be achieved by dynamically assembling neural modules, which have been discussed in dynamic neural networks \cite{dynamic_neural_network_survey} and neurosymbolic programming \cite{Neuro-symbolic}. 
Dynamic neural networks are a type of DNNs that support flexible inference based on the difficulty of input. When the input is easy, dynamic neural networks can reduce the computation by skipping a set of blocks \cite{skipnet, blockdrop} or exiting from the middle layers \cite{early_exit, hapi, adaptive_inference, flexdnn}.
Although different subnets (a subnet is a path in the dynamic NN) have different computational load, they cannot be customized for different tasks, and the input-dependent model dynamicality is also not suitable for edge environments.
Neurosymbolic programming demonstrates the possibility to solve different tasks with a same set of neural modules by semantically combining them. Specifically, the model to solve a complex task can be written as a program that invokes smaller function modules.
However, existing neurosymbolic programming approaches are short of flexibility and resource-oriented customization ability, because the modules are usually static and explicitly defined.

However, it remains challenging to optimize the model customization process for tasks and devices at the same time. The task-oriented customization and the resource-oriented customization are somewhat conflicting. To achieve fast task customization, the model is desired to have cross-task generalization ability, which typically requires a unified pretrained model with large capability, while deploying the model to diverse resource-limited edge devices requires non-uniform lightweight models.




\section{\name Design}
\label{sec:approach}

\textbf{Problem definition.} Formally, the goal of edge DNN customization is to generate a model $f_{\hat{\alpha},\hat{\theta}}$ with architecture $\hat{\alpha}$ and parameters $\hat{\theta}$ that can accurately handle the edge task while satisfying the performance constraints. \ie

\begin{equation}
\begin{split}
\hat{\alpha}, \hat{\theta} =\: & \arg \min_{\alpha, \theta}\: {L(f_{\alpha,\theta}(X^{e}), Y^{e})} \\
& s.t.\: mem(\alpha) < MEM^{e}\: and\: lat(\alpha) < LAT^{e}
\end{split}
\end{equation}
where $X^{e}$ and $Y^{e}$ are the input and output of edge task data samples, $L$ is the prediction loss, and $MEM^{e}$ and $LAT^{e}$ are the memory and latency limits at the edge environment.

Specifically, we focus on generating models for the tasks in a combinatorial space, \ie each task is described as a combination of several properties (\eg color, status, entity, etc.) and the properties are shared across different tasks (\eg red truck, white cat, etc.). Developers can flexibly define the task space according to their target edge scenarios.

Our work takes inspiration from the modularity and dynamic composability of neural networks and generative AI.
Our vision is to create an one-for-all generative system in which different edge-specific DNN models can be directly generated by configuring and assembling predefined neural modules, as illustrated in Figure~\ref{fig:illustration}.

Realizing this vision is challenging because (1) it is hard to design and develop the modules that can be assembled to do different tasks and fulfill different resource constraints (2) even if such modules are created, generating a tailored model for a specific edge scenario may still be difficult due to the huge combinational space of model candidates.

We attempt to address these challenges with an end-to-end approach named \name. We highlight that \name is a first-of-its-kind generative paradigm of model customization for diverse edge scenarios, which enables rapid training-free model generation for different tasks and resource constraints.

\subsection{Overview}

\begin{figure*}
    \centering
    \includegraphics[width=0.8\textwidth]{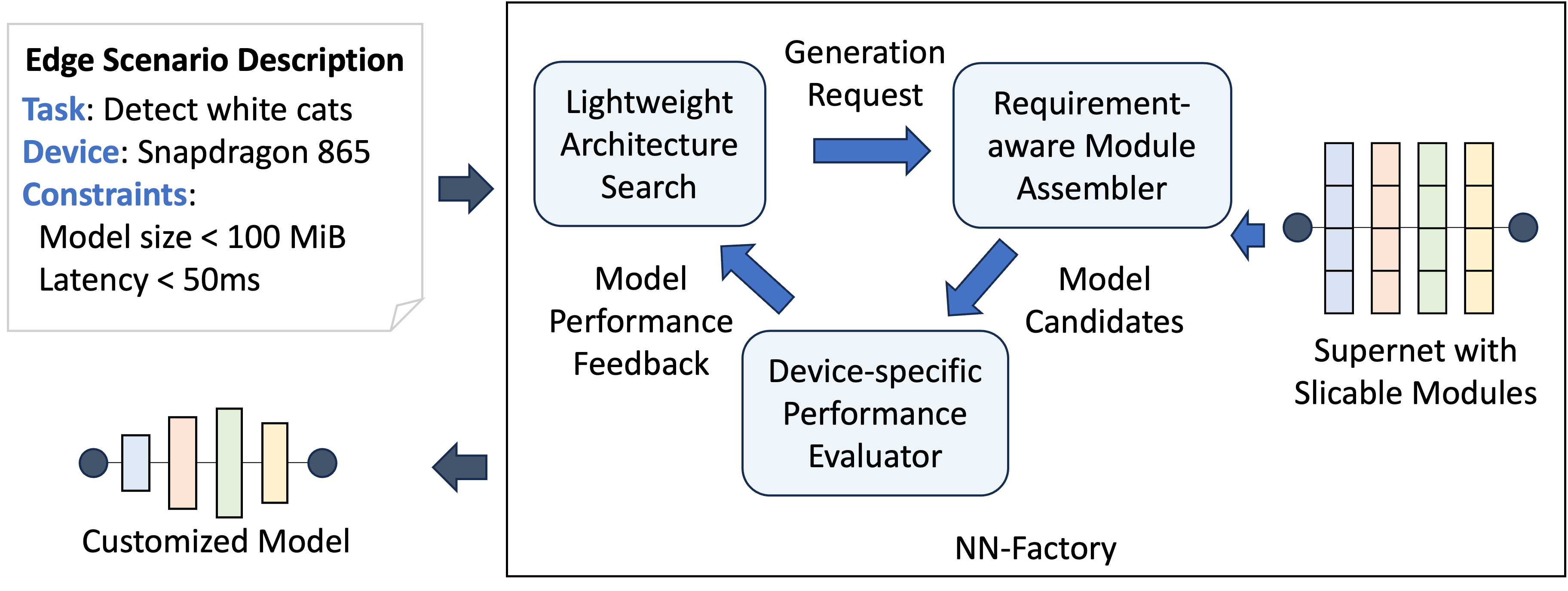}
    \caption{Architecture overview of \name.}
    \label{fig:architecture}
\end{figure*}

The main idea of \name is to jointly train a set of neural modules and a module assembling strategy generator (\ie assembler) in a task- and sparsity-grounded manner.
Specifically, the modules are sliced from a pretrained neural network (supernet), which avoids the cumbersome development of individual modules with the cost of limited interpretability. 
Meanwhile, by learning the assembler during training with many different tasks and sparsity requirements, \name can directly produce the model architecture for a new task and sparsity requirement with a single forwarding pass, significantly reducing the model search space.

Figure~\ref{fig:architecture} shows the overview of \name. It contains three main components, including a supernet with sliceable modules, a requirement-aware assembler, and a lightweight architecture search module. and a device-specific edge performance evaluator. The supernet contains the basic modules that can be flexibly combined to form customized models (subnets). The requirement-aware module assembler is a generative model that produces module assembling configurations based on the given task and sparsity requirements. Each generated configuration maps to a model candidate. The lightweight architecture search module finds the optimal model by iteratively searching and evaluating model candidates according to an edge-specific performance evaluator.

Given an edge scenario (described by the task, device type, and resource constraints), the model customization workflow of \name is consist of the following steps:

\begin{enumerate}
\item The lightweight architecture search module proposes a model generation request <\text{task}, \text{activation limit}>, where \text{activation limit} is the maximum ratio of activated modules in the model.
\item The requirement-aware assembler predicts a module assembling configuration based on the request. The configuration describes how to produce a candidate model by activating and assembling the modules in the supernet.
\item The device-specific performance evaluator evaluates the generated configuration (\ie the candidate model) against the edge resource constraints. If the constraints are satisfied and the memory/latency budget is fully utilized, then return the current candidate. Otherwise, go back to step (1) with a new generation request.
\end{enumerate}

The following subsections will introduce the main components in more detail.



\subsection{Supernet with Slicable Modules}
\label{sec:supernet}


The supernet is responsible for providing the basic neural modules that can be reassembled to fulfill different tasks and resource constraints.
Slicing an existing neural network for different function blocks for interpretability has been studied before \cite{zhang2020dynamic_slicing,zhang2022remos,zhang2023fedslice}, but the sliced modules are usually coarse-grained and difficult to reassemble. We decide to directly train a slicable supernet for better modular separation and composability.

We build the modular supernet by extending an existing backbone network. The backbone network extracts features from given inputs and make the final predictions. We are able to support common backbones based on Convolution Neural Networks (CNNs) and Transformers, such as ResNet \cite{resnet}, EifficentNet \cite{tan2019efficientnet}, and Vision Transformer \cite{dosovitskiy2020vit}.

\textbf{CNN Backbone.} First, we introduce how to convert a convolutional layer in the CNN architecture to slicable modules. Given a feature map $x$ as a input, the output of $l$-th convolutional layer is $O^{l}(x)$. In a conventional CNN, $O^{l}(x)$ is computed as:
\begin{equation}
O_i^{l}=\sigma(F_i^{l}*I^{l}(x)) \label{equation:1}
\end{equation}
where $O_i^{l}$ is the $i$-th channel of $O^{l}(x)$ , $F_i^{l}$ is the $i$-th filter, $\sigma(\cdot)$ denotes the element-wise nonlinear activation function and $*$ denotes convolution. the output feature map $O^{l}(x)$ is obtained by applying all the filters $F_i^{l}$ in the current layer to the input feature map $I^{l}(x)$.

In \name, we treat each convolutional filter as a module that can be conditionally activated. By activating different combinations of filters in a convolution layer, the resulting layers can deemed to have different functionalities.

\begin{figure}
    \centering
    \includegraphics[width=0.38\textwidth]{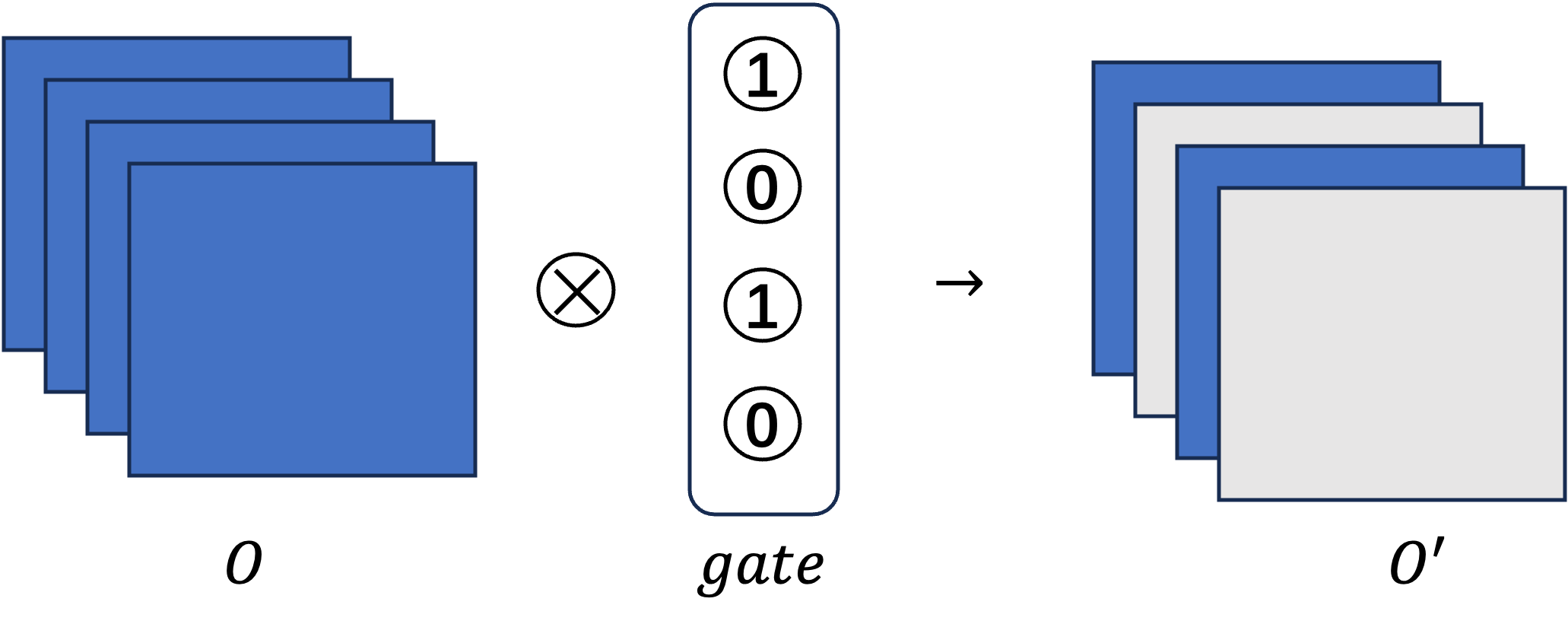}
    \vspace{-0.3cm}
    \caption{An illustration of modular convlutional layer with conditionally-activated convolution filters. $O$ represents the output feature map of a convolutional layer with four channels. Each element in the gate vector maps to a channel. By applying different gate vectors, different combinations of the filters are activated.}
    \label{fig:cnn}
\end{figure}

Based on such modular decomposition, we introduce a gate vector $g$ to control the activation of the modules. With the gate vector, the computation of feature map $O_i^{l}$ in Equation~\ref{equation:1} is reformulated as below: 
\begin{equation}
    O_i^{l}=\sigma(F_i^{l}*I^{l}(x)) \cdot g_i^{l}
\label{equation:gated_conv}
\end{equation}
where $g_i^{l}$ is the entry in $g$ corresponding to the $i$-th filter at layer $l$ and \textbf{0} is a \textbf{2-D} feature map with all its elements being 0, only when $g_i^{l}$ equals to 1, the $i$-th filter will be applied to $I^{l}$ to extract features.
%
Figure~\ref{fig:cnn} depicts the computation process as described above.

\begin{figure}
    \centering
    \includegraphics[width=\columnwidth]{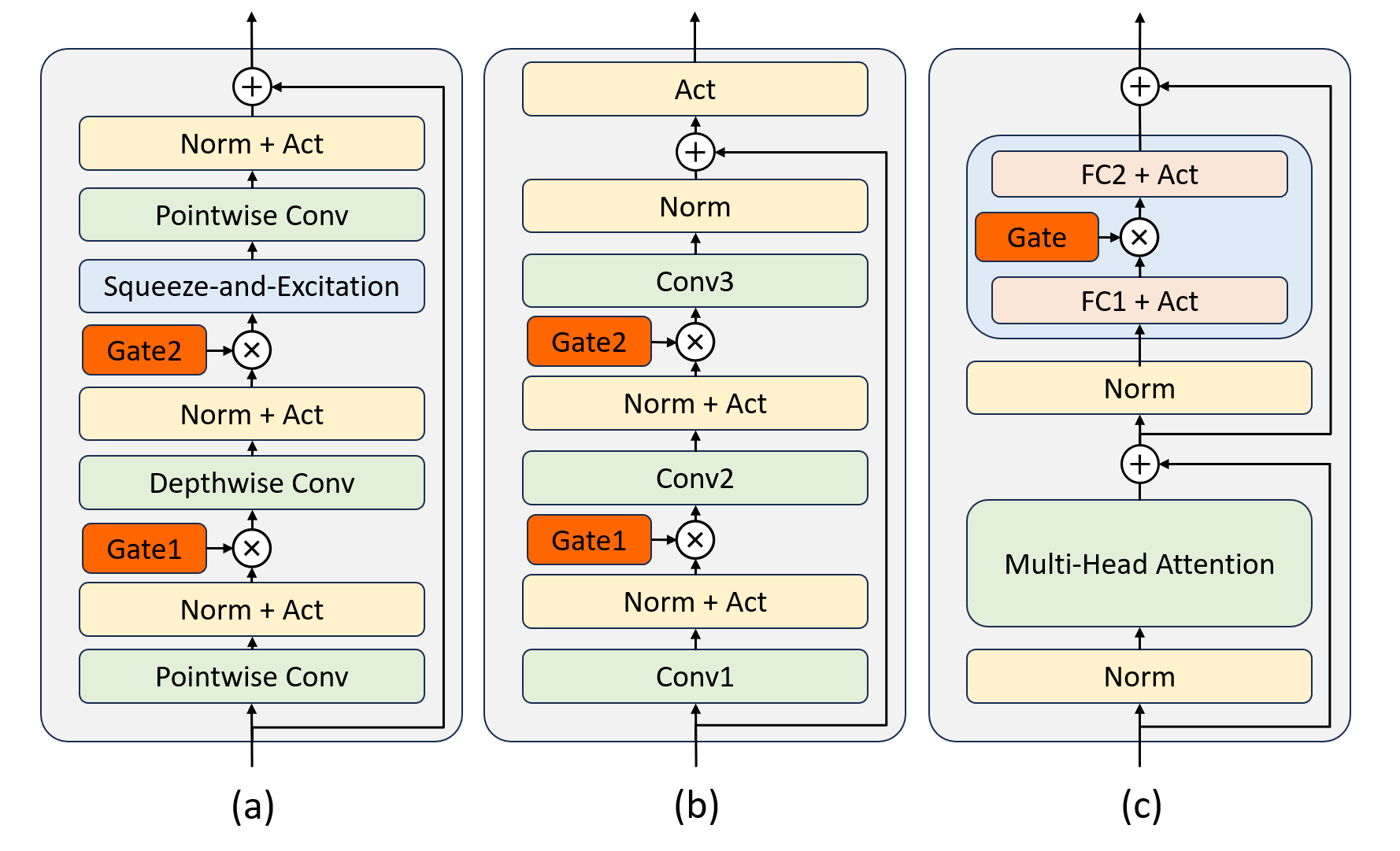}
    \vspace{-0.3cm}
    \caption{Gate integration for the different backbones. (a) EfficientNet and MobileNet. (b) ResNet. (c) ViT. }
    \label{fig:gate_backbones}
\end{figure}

The modular supernet is obtained by applying the modularization process to the main convolution layers in the CNN model. The batch normalization layer after each convolution layer is sliced in a channel-wise manner and controlled by the same gate vector. It is worth noting that, in modern deep CNN architectures such as ResNet, EfficientNet, and MobileNet, the convolutional layers are organized as several basic blocks. We do not convert the convolutional layers at the end of each basic block in order to avoid shape conflicts with the residual connections.
Figure~\ref{fig:gate_backbones}(a) and (b) illustrate the common CNN basic blocks that are integrated with learnable module gates.

\textbf{Transformer backbone.}
We also support generating modular supernets from Transformer backbones. The main components of Transformer include self-attention layers and feed-forward networks (FFNs).
According to recent studies \cite{dai2021knowledge}, the FFN components store various factual knowledge learned from data.
A FFN is a two-layer fully connected networks, which process an input representation $x \in \mathbb{R}^{dmodel}$ as:
\begin{equation}
\begin{aligned}
h &= xW_1 \\
F(x) &= \sigma(h)W_2
\label{equation:ffn}
\end{aligned}
\end{equation}
where $W_1 \in \mathbb{R}^{dmodel \times dff}$ and $W_2 \in \mathbb{R}^{dff \times dmodel}$ are the weight matrices.

\begin{figure}
    \centering
    \includegraphics[width=0.36\textwidth]{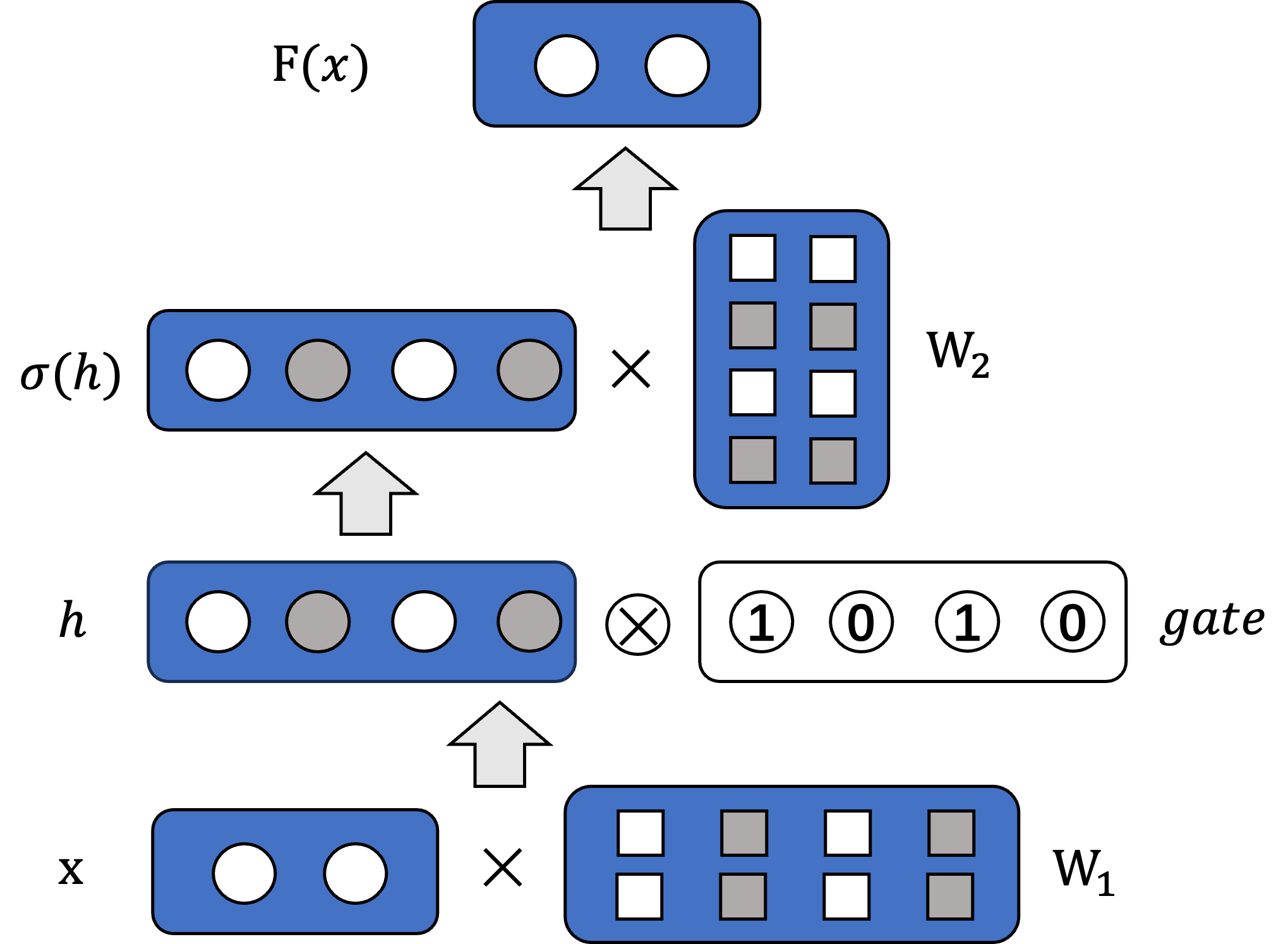}
    \caption{An illustration of modular FFN layer in Transformer backbones. The intermediate feature $h$ of FFN is conditionally activated along the hidden dimension. Each element in the gate vector maps to a element in $h$. According to the activation of $h$, the corresponding dimensions in $W_1$ and $W_2$ can be selectively pruned.}
    \label{fig:transformer}
\end{figure}

The Transformer architecture is not modular by design, but it can be converted into an equivalent Mixture-of-Experts (MoE) model \cite{MoE,zhang2021moefication}, in which each expert can be regarded as a conditionally-activated function module.
Inspired by this idea, we propose to modularize the FFN layers in Transformer backbones and reformulate the computation of $F(x)$ in Equation~\ref{equation:ffn} as below: 
\begin{equation}
\begin{aligned}
h &= xW_1 \\
h^{'} &= h \cdot g_{FFN} \\
F(x) &= \sigma(h^{'})W_2
\end{aligned}
\end{equation}
where $g_{FFN}$ represents the gate selection for the current FFN layer. If the $i$-th position in $g_{FFN}$ is 0, the corresponding position in $h'$ is set to 0 as well, indicating that the parameters in the corresponding parts of $W_1$ and $W_2$ are not activated.In Figure ~\ref{fig:transformer}, we demonstrate the parameters that remain inactive under the gate's selection.

Similarly, by applying the modularization process to all FFNs in a Transformer backbone, we can obtain a Transformer-based supernet for \name, an illustration with the ViT backbone is shown in Figure~\ref{fig:gate_backbones}(c).

\subsection{Requirement-aware Module Assembler}


\begin{figure}
    \centering
    \includegraphics[width=0.47\textwidth]{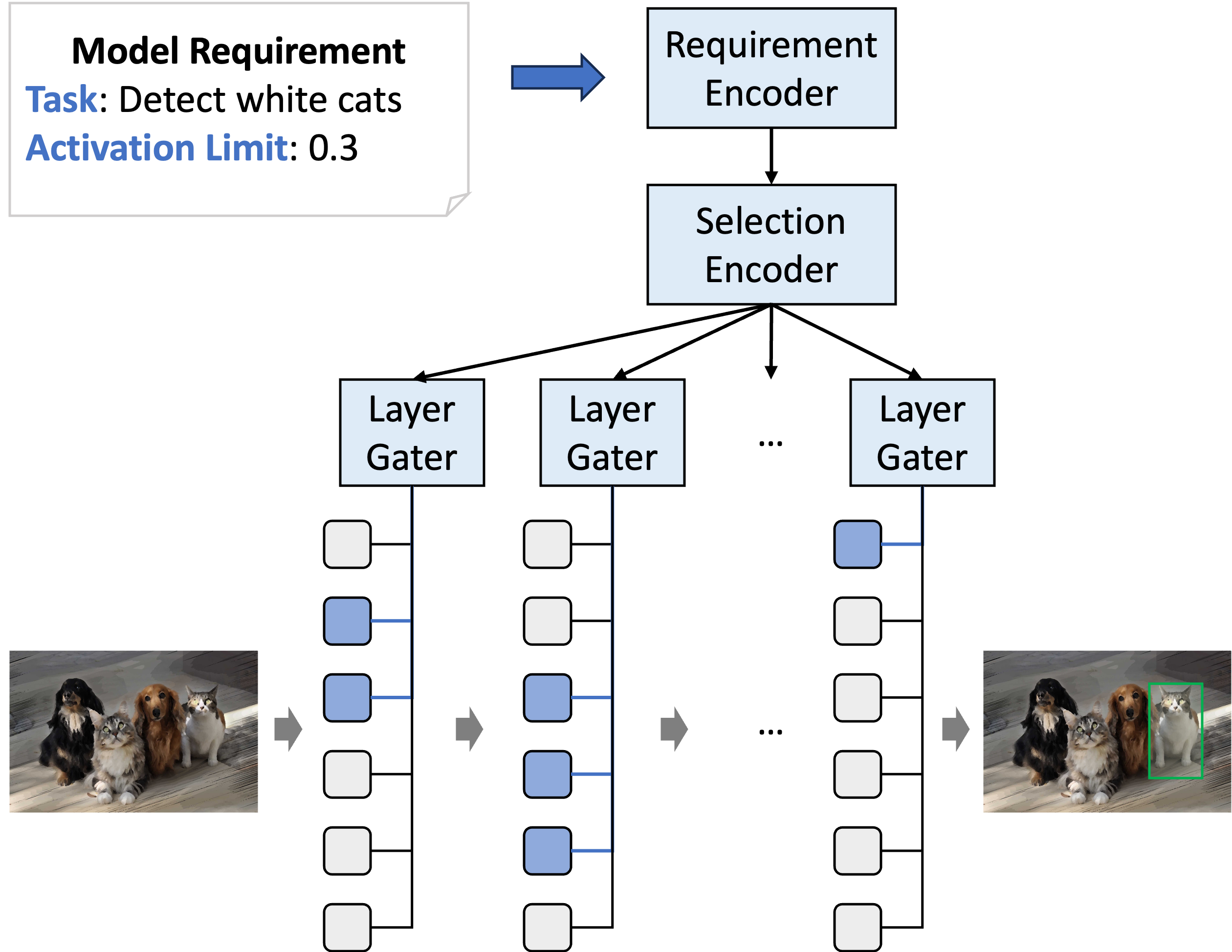}
    \caption{Architecture of the requirment-aware module assembler, integrated with the modular backbone.}
    \label{fig:assembler}
\end{figure}


The module assembler plays a vital role in \name - it generates gate vectors that controls the activations of neural modules in the supernet, such that the activated modules can be assembled to produce a candidate model.

The input of the module assembler is a model generation request, represented as a tuple <$task$, $activation\_limit$>. $task$ is a description of the target task in the edge scenario. As mentioned in Section~\ref{sec:approach}, a task can be represented as a combination of several properties. For example, the task `detect black dogs' is represented as $\{black, dog\}$ and `detect white cat' is $\{white, cat\}$. $activation\ limit$ is a sparsity requirement that regulates the ratio of activated modules in the generated model. A lower limit will encourage the assembler to generate smaller candidate models.

The output of the assembler is the gate vectors $g$ as described in Section~\ref{sec:supernet}. Each gate vector $g_l$ corresponds to a modular layer $l$ in the supernet and determines the activations of the modules in the layer.

The network architecture of the assembler is shown in Figure~\ref{fig:assembler}. It consists of three main components, including a requirement encoder, a selection encoder, and a list of layer gaters. The requirement encoder transforms the description of task and sparsity requirement $activation\_limit$ into an embedding.
We use the one-hot encoding for the $task$ component, denoted by $enc_{task}$. Each bit in $enc_{task}$ represents the presence of a specific property. For example, in the task `detect black dog', the elements for `black' and `dog' are set to one and others are set to zero in the task encoding. Regarding the $activation\_limit$ component, we utilize Positional Encoding \cite{transformer}  to encode the limit ratio, denoted by $enc_{limit}$. The two encodings $enc_{task}$ and $enc_{limit}$ are concatenated together as the requirement encoding $enc_{req}$.

The selection encoder convert the task encoding $enc_{req}$ to a intermediate representation $enc_{sel}$ that contains the knowledge about the whole-model gate selections. We use a fully-connected layer followed by a batch normalization (BN) and a ReLU unit to do this conversion.
The global selection encoding $enc_{sel}$ is then fed into each layer gater to compute the activations for each layer.
The layer gater is also a fully-connected layer followed by BN and ReLU unit. Its output value $weight^l \in \mathbb{R}^{D_l}$ represents the weights of modules in the layer $l$, where $D_l$ is the number of modules. The gate selections for the layer $g^l$ is obtained by discretizing $weight^l$ with a threshold (\ie $g^l = weight^l > threshold$). We set $threshold = 0.5$ by default in our implementation.




\subsection{Joint Training and Separate Deployment}


\textbf{Supernet-Assembler Joint Training.}
The modular supernet and assembler are closely related - the assembler produces the activations of modules, and the activations represent the candidate models with the supernet.
Therefore, we train them jointly to make them coherently work together.

The training is conducted with a set of training tasks and labeled data belong to each task.
Each sample can be represented as a tuple $\{task, activation\_limit, x, \hat{y}\}$, where $x$ and $\hat{y}$ are the input and label. All samples are mixed together and shuffled during training.

In each forward pass, we compute the gate selections $g$ by feeding $task$ and $activation\_limit$ into the assembler, obtain the subnet $M_g$ with the gate selections $g$, and get the prediction $y$ by feeding $x$ into the subnet $M_g$.
The loss is computed by checking the prediction $y$ and the gate selections $g$, \ie
\begin{equation}
L=TL(y, \hat{y}) + \lambda\ GL(g, activation\_limit)
\end{equation}
$TL(y, \hat{y})$ is the loss for training task, which encourages the generated subnet to produce correct predictions based on the underlying task.
$GL(\cdot)$ is the loss for the gate selections, which encourages the assembler to generate subnets that satisfy the sparsity requirement. If the ratio of ones in $g$ is lower than the given $activation\_limit$, $GL(\cdot)$ is disabled. Otherwise, $GL(\cdot)$ is the mean squared error (MSE) between the ratio of ones in $g$ and $activation\_limit$. $\lambda$ is a hyperparameter to balance the two objectives, which is set to 100 by default.



Note that the gate selection loss $GL(\cdot)$ is computed with the discrete binary gate selections $g$, but the actual output of the assembler is the continuous gate weights $weight$, which may lead to the difficulty of error backpropagation.
To address this issue, we employ a method called \emph{Improved SemHash} \cite{kaiser2018fast,kaiser2018discrete} to do the trick. The computation is performed as follows: 
\begin{gather*}
    g_\alpha= 1(weight>0)  \\
    g_\beta  = \max\left(0, \min\left(1, 1.2\sigma(weight) - 0.1\right)\right) 
\end{gather*}
Here, $g_\alpha$ is a binary vector, while $g_\beta$ is a real-valued gate vector with all entries falling in the interval [0.0, 1.0]. $g_\alpha$ has the desirable binary property that we want to use in our training, but the gradient of $g_\alpha$ is zero for most values of $g$. In contrast, the gradient of $g_\beta$ is well defined, but $g_\beta$ is not a binary vector.  In the forward pass during training, we randomly use $g$ = $g_\alpha$ for half of the samples and use $g$ = $g_\beta$ for the rest. When $g_\alpha$ is used, we follow the solution in \cite{kaiser2018fast,kaiser2018discrete} and define the gradient of $g_\alpha$ to be the same as the gradient of $g_\beta$ in the backward propagation. For evaluation and inference, we always use the discrete gates. 

\textbf{Separate Deployment of Subnet.} After the joint pretraining, we are able to generate the gate selections based on different task and sparsity requirements. Once the gate selections are determined, the assembler network and the deactivated modules in the modular supernet are no longer useful. Therefore, when deploying the model, we only have to transmit and deploy the subnet assembled with the active modules. For example, with CNN-based supernet, we can prune the inactive filters based on gate selection. By doing so, the resulting subnet would have significantly reduced model size, memory usage, and latency. For Transformer-based supernet, we can achieve similar results by pruning the corresponding parameters in $W1$ and $W2$ based on gate selection. 
The deployed model does not require any further architecture change or parameter training.

\subsection{Lightweight Architecture Search}

The jointly-trained modules and assembler enable us to efficiently produce candidate models with different tasks and sparsity levels.
Based on such an ability, we further introduce a lightweight architecture search strategy to find the optimal model for each edge scenario.

The architecture search process is guided by a device-specific performance evaluator, which takes a candidate model as the input and tells whether the model satisfies the resource constraints (\ie the inference latency is small than the latency budget and the memory cost is small than the memory budget).
We can directly evaluate the performance on the target edge devices, following the on-device subnet selection practice of AdaptiveNet \cite{wen2023adaptivenet}.

\begin{figure}
    \centering
    \begin{subfigure}[b]{0.47\linewidth}
      \centering
      \includegraphics[width=\linewidth]{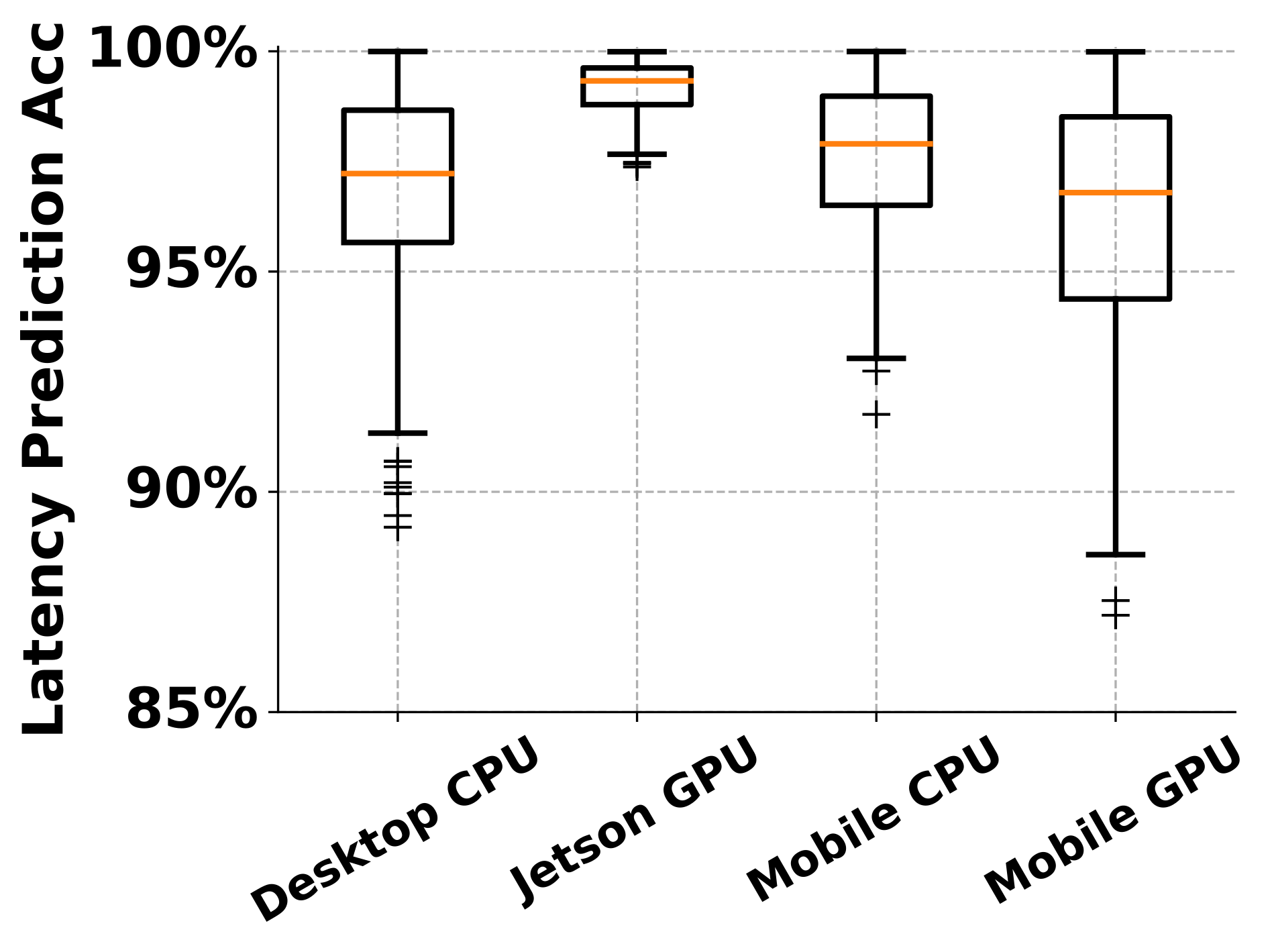}
      \caption{Latency}
      \label{fig:perf_pred_acc:latency}
    \end{subfigure}
    \begin{subfigure}[b]{0.47\linewidth}
      \centering
      \includegraphics[width=\linewidth]{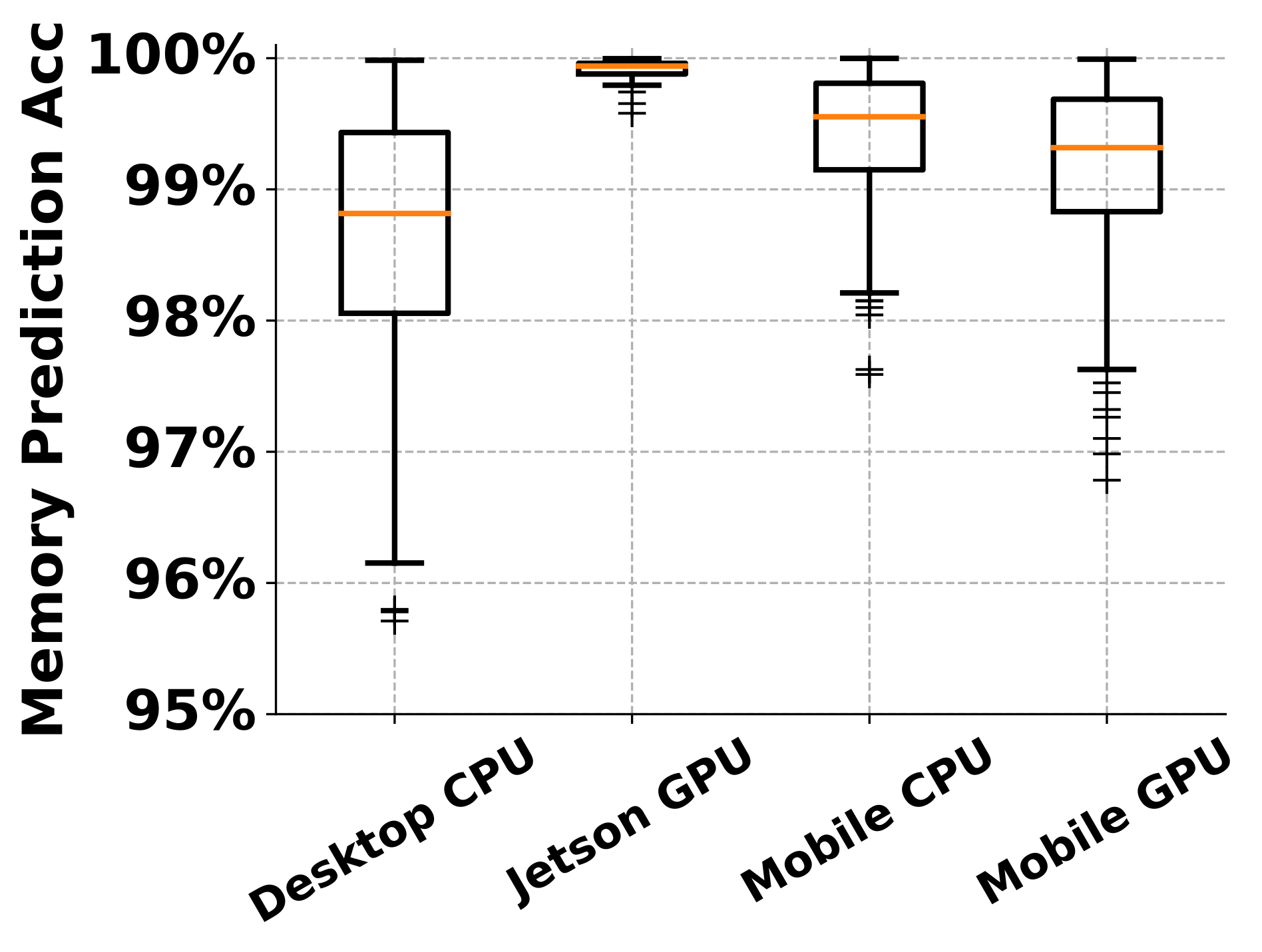}
      \caption{Memory}
      \label{fig:perf_pred_acc:memory}
    \end{subfigure}
    \caption{Performance prediction accuracy of \name generated models on four edge devices.}
    \label{fig:perf_pred_acc}
\end{figure}

Alternatively, when deploying the supernet and assembler to the edge device is uneasy, we can build a performance predictor to evaluate the candidate models with profiling data collected from the edge devices.
Such a profiling-based device performance modeling is a common practice in mobile/edge computing \cite{legodnn}, while here we make some reasonable simplifications based on the our modular design.
Specifically, we take a layer-wise performance profiling and modeling approach, in which the total latency of the model equals the sum of all layers (with a static bias), and the latency of each layer is dependent on the gate vector predicted by the module assembler.
The modeling of memory consumption is similar, which is determined by the static memory consumption (the model parameters) and peak memory at runtime (the largest intermediate feature).
To build these predictors, we generate a set of (2000 in our experiments) random gate vectors, obtain the corresponding subnets, measure the performance metrics of these subnets on target edge devices, and use the collected data to train the performance predictors with linear regression.
The predictor is highly accurate, with both the latency prediction accuracy and the memory prediction accuracy higher than 96\% on four typical edge devices, as shown in Figure~\ref{fig:perf_pred_acc}. We use the predictor by default in \name.

\begin{algorithm}
    \caption{Lightweight Architecture Search}
    \label{algo:search}
    \begin{algorithmic}[1]
        \Require Target edge scenario $e$, task description $task^e$, latency requirement ${LAT}^{e}$, memory requirement ${MEM}^{e}$, edge-specific performance evaluator $Evaluator^{e}$.
        \Ensure Customized edge model $M$
    
        \State $gate^{opt} \gets NULL$
        \For{$limit_{i}$ = 1\%; $limit_{i}$ < 1 ; $limit_{i}$ += step}
            \State $enc_i \gets Requirement\_Encoding(task^e,limit_{i})$
            \State $gate_i \gets Assembler(enc_i)$
            \State $lat_{enc},mem_{enc} \gets Evaluator^e(gate_i)$
            \If{$lat_{enc}>{LAT}^{e}$ or $mem_{enc}>{MEM}^{e}$}
                \State break
            \EndIf
            \State $gate^{opt} \gets gate_i$
        \EndFor
        \State \textbf{return} the subnet sliced from supernet with $gate^{opt}$
    \end{algorithmic}
\end{algorithm}

Based on the performance evaluator, we are able to analyze the latency and memory of candidate models. As the more modules activated usually lead to higher accuracy (see Section~\ref{eval:generation-performance}), the optimal model for an edge scenario is the one with the highest module activation ratio while satisfying memory and latency requirements.
Algorithm~\ref{algo:search} shows our lightweight architecture search strategy.
We start from the lowest module activation limit $limit_{i}=1\%$, and iteratively raise the limit. For each activation limit, we generate the gate selections using the assembler network, and obtain the latency and memory of related to the gate selections.
The last candidate gate selections that meet the edge environment constraints are identified and returned.
Finally, we generate the subnet with the result gate selections, which can be directly deployed to the edge without any further processing.

    \section{Implementation}
\label{sec:implementation}


We implement our method using Python. The training part is based on PyTorch. The models are generated with PyTorch and deployed to edge devices using TensorFlow Lite framework for mobile and PyTorch for Desktop and Jeston


\textbf{Architecture and Training Details.} In the layer gaters of the assembler, we employ distinct batch normalization layers for the selection encoding of different layers, this may influence the quality of the generated model. In order to enhance training stability and improve the quality of the generated model, we augmented the training task set with additional tasks to incorporate more combinations of task properties, which can enhance the model's understanding of our task encoding. In order to incorporate broader sparsity requirements during training without compromising the model quality, we employ multiple layer gaters to enhance the capacity of the model, drawing on the principles of the Mixture of Experts (MoE), each gater converts the selection encoding into gate selections. Additionally, a gating network is introduced to determine the weighting of outputs from each gater at each layer.

\section{Evaluation}
\label{sec:experiment}


We conduct experiments to answer the following questions:
(1) Is \name able to generate edge-specific models? How is the quality of the generated models?
(2) How much is the overhead of \name?
(3) How well can the model generation ability of \name generalize to unseen edge scenarios?

\subsection{Experimental Setup}
\label{eval:setup}

\textbf{Tasks and Datasets.}
We evaluated the performance of \name on two model customization settings.
\begin{itemize}
\item \textbf{Numeric Visual Question Answering (NumVQA).} This is a simple setting to analyze the performance of task- and resource-specific model generation. The task is to answer a yes-or-no question (\eg ``Are there two even numbers?'') based on an input image containing four digits. We formulated approximately 60 questions and synthesized images using the MNIST dataset \cite{lecun1998mnist}. The performance of generated models was measured by the classification accuracy.
\item \textbf{Attributed Object Detection (AttrOD).} This is a more practical setting, in which each task is to detect objects with specific attributes in an image and predict the object bounding boxes and categories. We utilized a vision question answering model \cite{kim2021vilt} to annotate the object colors in the COCO2017 \cite{coco} dataset and merged identified colors to form target attributes. We selected 5 attributes (1-white, 2-bronze, 3-charcoal, 4-crimson, 5-chartreuse) from 4 categories and combined them to construct our training tasks. The total number of tasks was approximately 140, and each task has a varying number of samples, ranging from hundreds to tens of thousands.
The performance of detection models was measured by mean average precision over Intersection over Union threshold 0.5 (mAP@0.5).
\end{itemize}

\textbf{Model Backbones.}
We considered the common CNN and Transformer backbones in this experiment, including ResNet \cite{resnet}, ViT \cite{dosovitskiy2020vit},  for the NumVQA setting and EfficientDet \cite{efficientdet} for the AttrOD setting.

\textbf{Baselines.}
We compared \name with two conventional model customization approaches:
\begin{enumerate}
  \item \textbf{Retrain} - We fixed the model architecture and retrained it with standard supervised learning method for each edge scenario.
  \item \textbf{Prune\&Tune} - We trained a unified model. Given an edge scenario, we pruned and fine-tuned the pretrained model to match edge requirements with a SOTA pruning method~\cite{pruner}.
\end{enumerate}
Both of them require training with edge-specific data. We used the same backbones with these baselines and trained them until convergence. We did not include other training-free model generation/scaling methods \cite{oneforall,legodnn,nestdnn,wen2023adaptivenet} since they don't support task-oriented customization.

\textbf{Edge Environments.} 
We considered three edge devices including an Android Smartphone (Xiaomi 12) with Snapdragon® 8 Gen 1 processor and 12GB memory, a Jetson AGX Xavier with 32 GB memory, and a desktop computer with 12th Gen Intel® Core™ i9-12900K $\times$ 24 Processor with 64GB memory.
The batch sizes were all set to 1 on the three devices to simulate real workloads.
We used different latency budgets to simulate intra-device hardware diversity.

\subsection{Model Generation Quality}
\label{eval:generation-performance}
We conducted an evaluation of the models generated by \name on NumVQA and AttrOD, followed by a comprehensive analysis of their quality.

\begin{table}[]
\centering
\caption{The quality of generated models on NumVQA.}
\label{tab:Number}
\resizebox{\columnwidth}{!}{
\begin{tabular}{cc|cc|cc}
\toprule

\multicolumn{2}{c|}{\textbf{Generation Request}} & \multicolumn{2}{c|}{\textbf{ResNet}} & \multicolumn{2}{c}{\textbf{ViT}}\\

\textbf{Task} & \textbf{Act. Limit}  &   \textbf{Acc} & \textbf{Act. Ratio}  &   \textbf{Acc} & \textbf{Act. Ratio}   \\

\midrule
Has a number 0  & 3\% & 99.7\% & 3.0\% & 97.8\% & 3.0\% \\
Has a number 0  & 5\% & 99.8\% & 3.7\% & 97.6\% & 4.0\% \\
Has a number 0  & 10\% & 99.9\% & 3.8\% & 97.8\% & 5.1\% \\
Only two number 1  & 3\% & 99.3\% & 3.0\% &  96.7\% & 3.0\% \\
Only three number 2  & 3\% & 99.4\% & 3.0\% &  86.7\% & 3.0\% \\
Only four number 5  & 3\% & 99.1\% & 3.0\%&  95.5\% & 3.0\%  \\
Only one number 0  & 5\% & 99.8\% & 3.7\% & 94.1\% & 5.0\% \\
Only three number 0  & 5\% & 99.4\% & 3.8\% &  94.8\% & 5.0\% \\
Only one number 3  & 5\% & 99.8\% & 3.7\% &  87.7\% & 5.0\% \\
Only two odd numbers  & 3\% & 99.3\% & 3.0\%&  94.3\% & 3.0\%  \\
Only two odd numbers  & 5\% & 99.4\% & 4.0\%&  94.3\% & 3.8\%  \\
\bottomrule
\end{tabular}
}
\end{table}

\begin{table*}[ht]
\centering
\caption{The quality of models generated by \name and baselines for different edge scenarios in the AttrOD setting. The `task' column is the target object followed by the desired attributes (\eg black, yellow). The `Lat' and `Mem' columns are the relative latency and memory as compared with the requirements ($LAT^{req}$ and $MEM^{req}$).}
\label{tab:detection}
\resizebox{\textwidth}{!}{
\begin{tabular}{cccc|ccc|ccc|ccc}
\toprule
\multicolumn{4}{c|}{\textbf{Edge Scenarios}} & \multicolumn{3}{c|}{\textbf{\name}} & \multicolumn{3}{c|}{\textbf{Retrain}} & \multicolumn{3}{c}{\textbf{Prune\&Tune}} \\
\text{Task} & \text{Device} & \text{$LAT^{req}$} & \text{$MEM^{req}$} & $\Delta$Lat & $\Delta$Mem & mAP & $\Delta$Lat & $\Delta$Mem & mAP & $\Delta$Lat & $\Delta$Mem & mAP \\
\midrule
Bicycle \{3,4\} & Desktop & 55ms  &  0.30GB & \textcolor{blue}{-0.11} & \textcolor{blue}{-0.03} & 0.31  &   \textcolor{red}{+156.77} &  \textcolor{red}{+0.38} & \textbf{0.35} & \textcolor{blue}{-0.33} & \textcolor{blue}{-0.23} & 0.34 \\
Person  \{2\} & Desktop  & 70ms  &  0.35GB & \textcolor{blue}{-4.55}  & \textcolor{blue}{-0.05}  & \textbf{0.27} &  \textcolor{red}{+141.77} & \textcolor{red}{+0.33} & 0.24 & \textcolor{blue}{-1.05} & \textcolor{blue}{-0.22} & 0.22 \\
Car \{3\} & Desktop  & 85ms  & 0.40GB & \textcolor{blue}{-2.96} &\textcolor{blue}{-0.01} & 0.38 & \textcolor{red}{+126.77} & \textcolor{red}{+0.28}  & \textbf{0.42} & \textcolor{blue}{-1.77} & \textcolor{blue}{-0.22} & 0.41 \\
Motorcycle \{1\} & Desktop  &  100ms  & 0.45GB &  \textcolor{blue}{-0.05}  & \textcolor{blue}{-0.02} & \textbf{0.45} & \textcolor{red}{+111.77} & \textcolor{red}{+0.23} & 0.23 & \textcolor{blue}{-0.70} & \textcolor{blue}{-0.21} & 0.24 \\
Motorcycle \{2,3,4\} & Desktop & 115ms & 0.50GB & \textcolor{blue}{-2.43} & \textcolor{blue}{-0.04} &  0.43 & \textcolor{red}{+96.77} & \textcolor{red}{+0.18} & 0.44 & \textcolor{blue}{-1.42} & \textcolor{blue}{-0.20} & \textbf{0.45} \\
Bicycle \{3,4\} & Mobile & 350ms & 5GB & \textcolor{blue}{-47.36} & \textcolor{blue}{-0.24} & 0.31 & \textcolor{red}{+625.28} & \textcolor{red}{+19.97} & \textbf{0.35} & \textcolor{blue}{-54.4} & \textcolor{blue}{-0.21} & 0.34 \\
Motorcycle \{5\} & Mobile  &450ms & 7GB & \textcolor{blue}{-47.54} & \textcolor{blue}{-0.07} &\textbf{0.53} & \textcolor{red}{+525.28} & \textcolor{red}{+17.97} & 0.19 & \textcolor{blue}{-86.43} & \textcolor{blue}{-0.21} & 0.19 \\
Car \{3\} & Mobile &550ms & 9GB & \textcolor{blue}{-25.92} &  \textcolor{blue}{-0.01} & 0.38 & \textcolor{red}{+425.28} & \textcolor{red}{+15.97} & \textbf{0.42} &  \textcolor{blue}{-119.46} & \textcolor{blue}{-0.18} & 0.41 \\
Person \{5\} & Jetson  & 15ms &  ---  & \textcolor{blue}{-0.61} & --- & \textbf{0.22} & \textcolor{red}{+42.98} & --- & 0.15  & \textcolor{blue}{-0.40} & --- & 0.07  \\
Person \{5\} & Jetson  & 20ms &  ---  & \textcolor{blue}{-0.81} & --- & \textbf{0.22} & \textcolor{red}{+37.98} & --- & 0.15 &\textcolor{blue}{-0.29}  & --- & 0.15 \\
Person \{5\} & Jetson  & 30ms &  ---  & \textcolor{blue}{-0.06}  & --- & \textbf{0.23}  & \textcolor{red}{+27.98} & --- & 0.15 & \textcolor{blue}{-0.08} & --- & 0.16 \\
\bottomrule
\end{tabular}
}
\end{table*}




We first assessed the effectiveness of the modules and assembler in the NumVQA setting. We feed different <$task$, $activation\ limit$> tuples to the \name assembler and let it generate models that meet the requirements. The results are presented in Table~\ref{tab:Number}. Overall, \name demonstrated exceptional accuracy (>99\% with ResNet backbone) across diverse tasks, while also ensuring that the generated models adhered to our specified activation limit. Due to the relatively small task scale, only a lower activation ratio is required. 

Next, we evaluated the end-to-end performance of \name on AttrOD. In Table ~\ref{tab:detection}, we employed ResNet50 as the backbone and showcased the quality of the generated models under varying latency and memory requirements on different devices, and compare them with baseline models. 
Overall, \name consistently delivers high-quality models that meet the specified criteria in all edge scenarios. Thanks to the performance-aware model architecture search, \name was able to full utilize the given budgets (either memory or latency). 
The Retrain baseline failed to meet the latency and memory requirements as it didn't adjust the model architectures for each scenario.

The accuracy of models generated by \name is close to the Retrain baseline and outperforms the Prune\&Tune baseline, although it doesn't require edge-specific training.
Meanwhile, it managed to achieve significantly higher mAP scores on some tasks (\eg Motorcycle \{1\}, Person \{5\}) that contain less training samples than others.
This was because the mixed training of tasks, each with different attribute combinations, enabled the model to achieve a more profound understanding of the tasks and, consequently, make more precise predictions. 
The accuracy of Prune\&Tune models is much lower than \name and Retrain due to the reduced model size.

\begin{figure}
    \centering
    \includegraphics[width=0.47\textwidth]{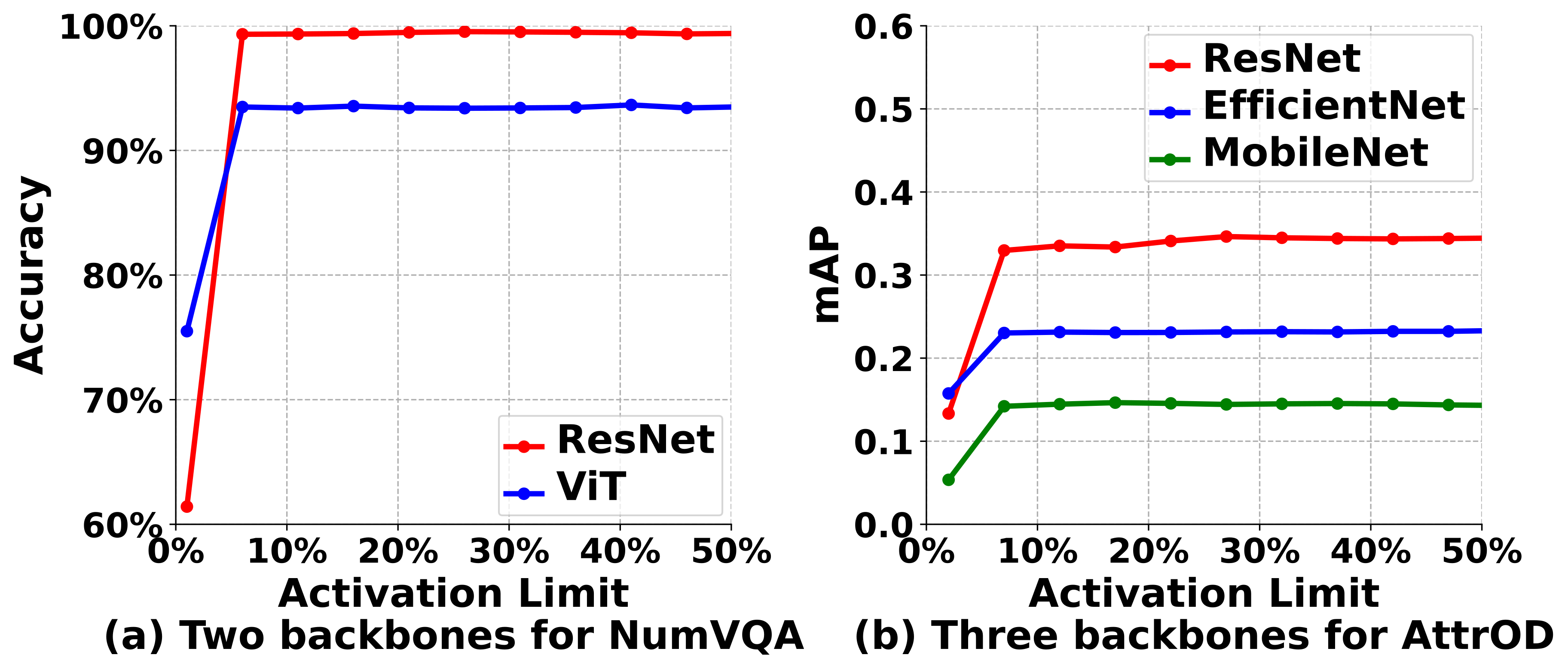}
    \vspace{-0.3cm}
    \caption{Accuracy of models generated by \name under different activation limits in NumVQA and AttrOD. The accuracy values are averaged from 10 random tasks.}
    \label{fig:TwoTask}
\end{figure}

We also explored the differences in the model quality of \name with different backbones. The results are displayed in Figure \ref{fig:TwoTask}. \name demonstrates consistent behavior across various supernet backbones, signifying its generalizablilty. However, it does exhibit distinct performance differences based on the chosen backbone. For example, the models generated by ViT-based \name and MobileNet-based \name exhahited lower accuracy. This was due to the inherent properties of backbone network, \eg poor sample efficiency of ViT and limited model capability of MobileNet.

\begin{figure}
    \centering
    \includegraphics[width=0.38\textwidth]{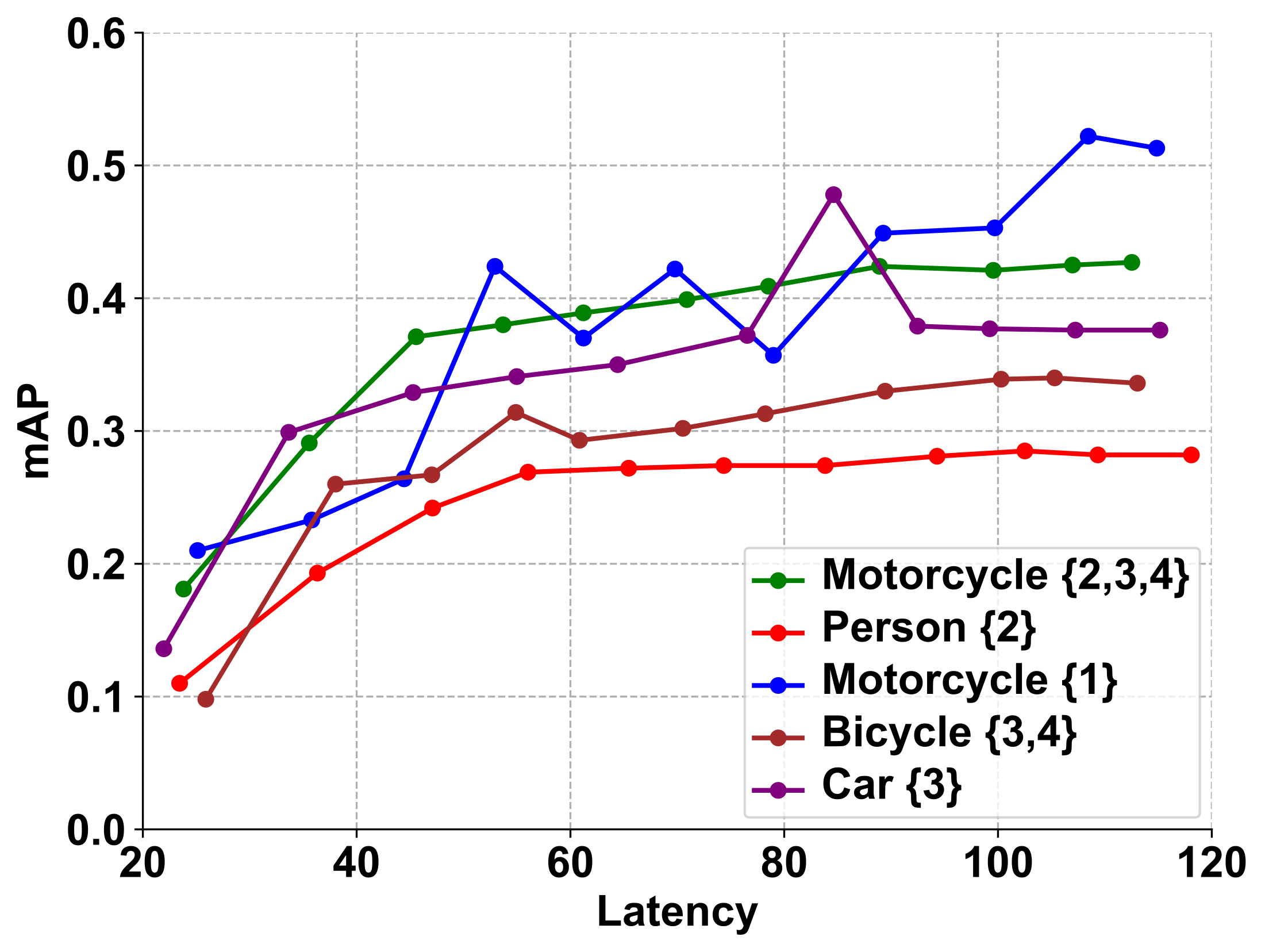}
    \vspace{-0.3cm}
    \caption{The trade-off between mAP (mean Average Precision) and latency for \name across different tasks of AttrOD.}
    \label{fig:tradeoff}
\end{figure}

\begin{figure}
    \centering
    \includegraphics[width=0.40\textwidth]{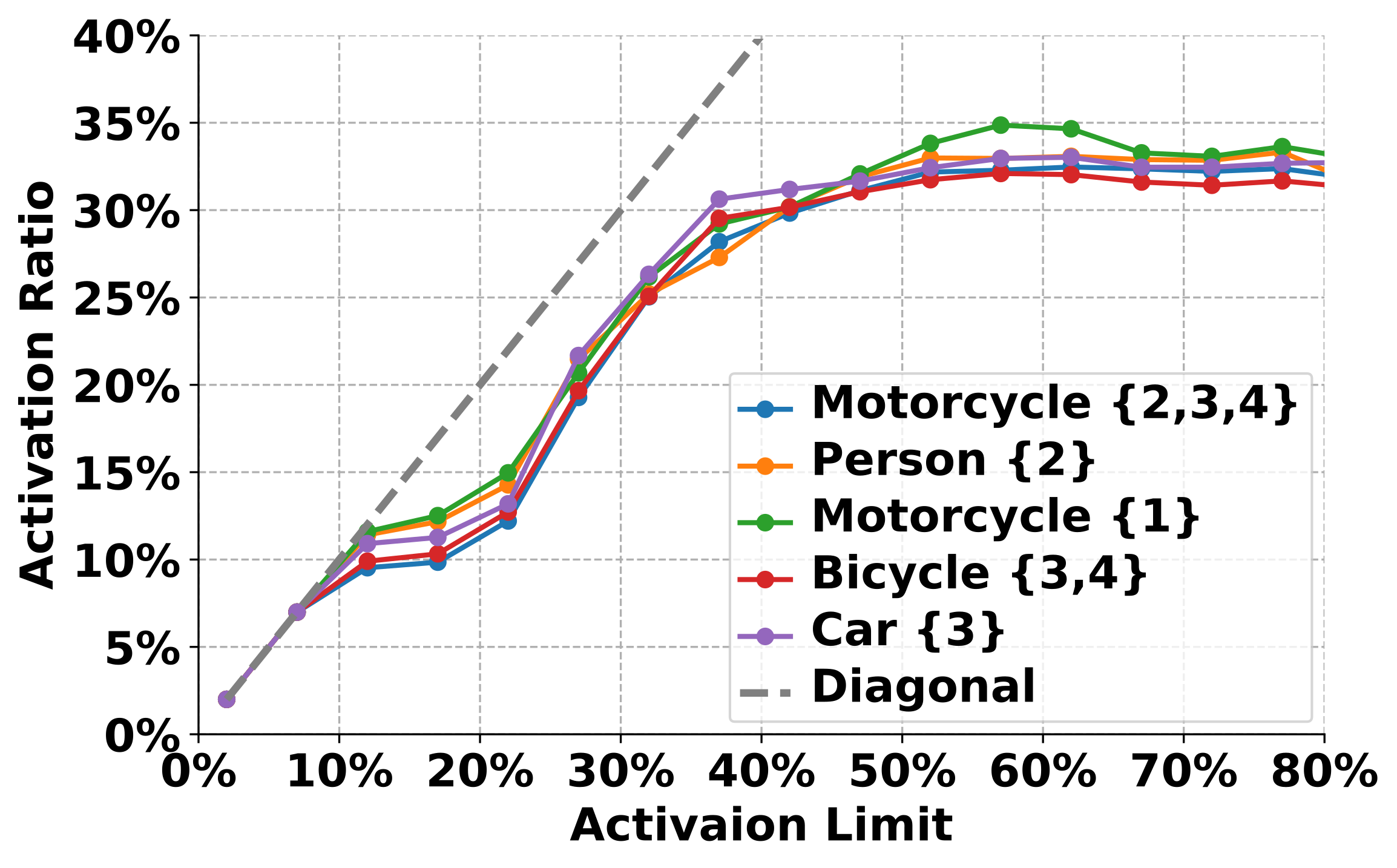}
    \vspace{-0.3cm}
    \caption{The relation between actual module activation ratio and specified activation limit.}
    \label{fig:TargetAndTrueUsage}
\end{figure}
Furthermore, We performed an analysis to evaluate how \name manages the trade-off between model quality and latency. The results are presented in Figure \ref{fig:tradeoff}, \name was capable of generating higher-quality models when given higher latency constraints. Figure \ref{fig:TargetAndTrueUsage} illustrates the correlation between the input activation limit during model generation and the activation ratio of the gate selections generated by \name. With a very low activation limit, generated activation ratio falls short of meeting the requirements. Consequently, it undergoes pruning based on importance, resulting in a pattern that closely aligns with the diagonal on the graph. With the increment of the activation limit, \name produced activation ratios that could satisfy the criteria. The ratios progressively improved with the increasing activation limits and eventually reached a stable state, because the stabilized ratios were already sufficient to produce the accurate predictions.

\subsection{Model Generation Efficiency}
\label{eval:generation-efficiency}


\begin{table}
    \centering
    \caption{The preparation time (once for all scenarios) and customization time (once for each edge scenario) of \name and baselines.}
    \label{tab:cost}
    \vspace{-0.3cm}
    \begin{tabular}{ccc} 
    \toprule 
    \textbf{Method} & \textbf{Preparation} & \textbf{Customization} \\
    \midrule 
    \name & 130 hours & 3.6s \\
    Retrain & -- & 66 hours \\
    Prune\&Tune & 66 hours & 4 hours \\
    \bottomrule 
    \end{tabular}
\end{table}

We conducted a comparative analysis of preparation and customization costs between \name and the baselines, and the results are presented in Table~\ref{tab:cost}.

\textbf{Customization Cost for Each Edge Scenario.}
The primary goal of \name is to reduce the model customization time. It took just 3.6 seconds on average for \name to generate a customized model for an edge scenario, demonstrating remarkable efficiency. This process includes about 12 rounds of searching, each taking 0.2 seconds, and model extraction, which takes 0.8 seconds. In contrast, traditional model customization methods incurred a staggering 4000-fold increase in time because they need to train the model for each edge scenario.


\begin{figure}
\centering
\begin{minipage}{0.48\linewidth}
    \centering
    \includegraphics[width=4.2cm]{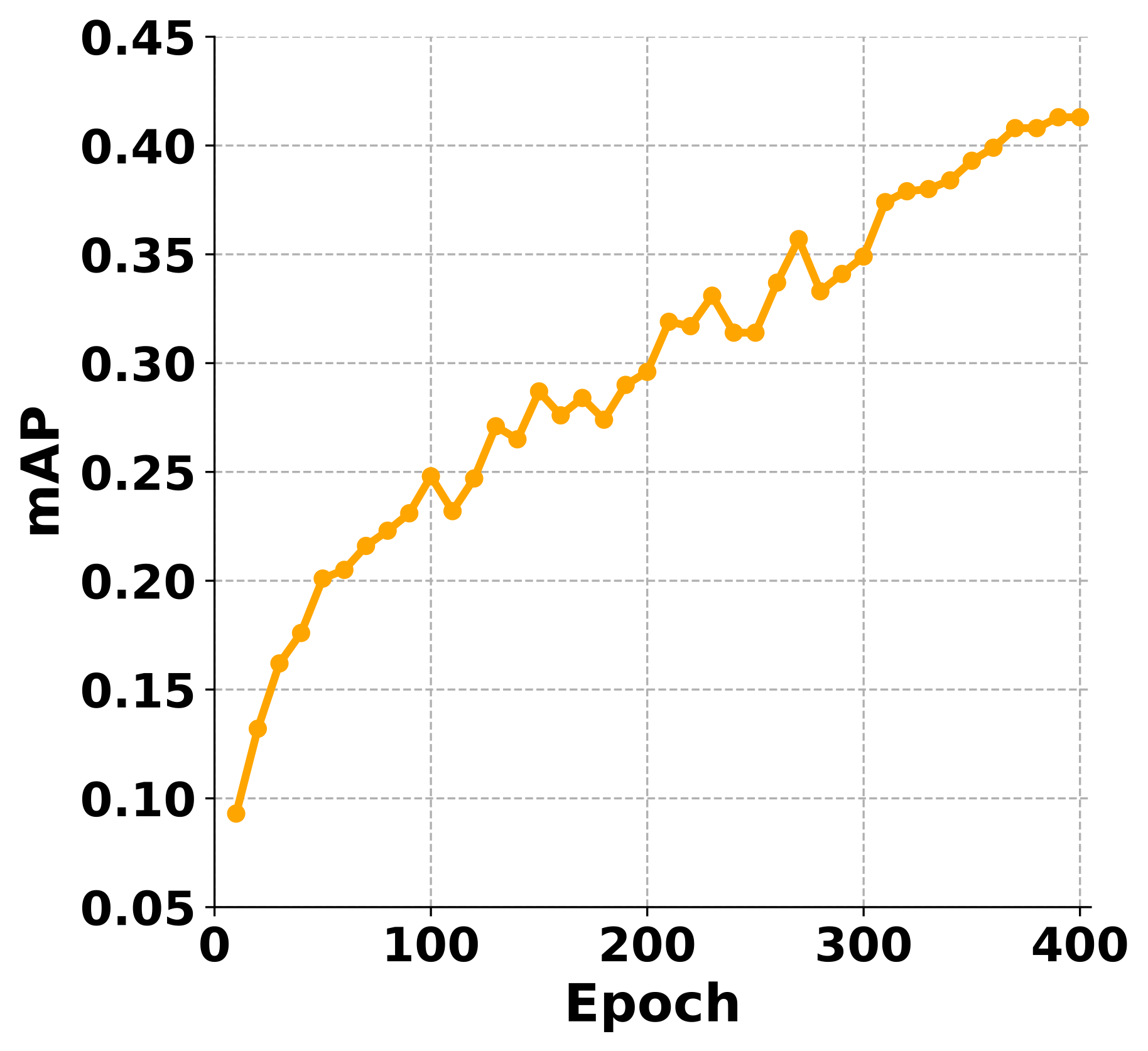}
    \vspace{-0.5cm}
    \caption{The average accuracy of generated models after training with different \#epochs.}
    \label{fig:EpochAcc}
\end{minipage}
\hfill
\begin{minipage}{0.48\linewidth}
    \centering
      \includegraphics[width=4.4cm]{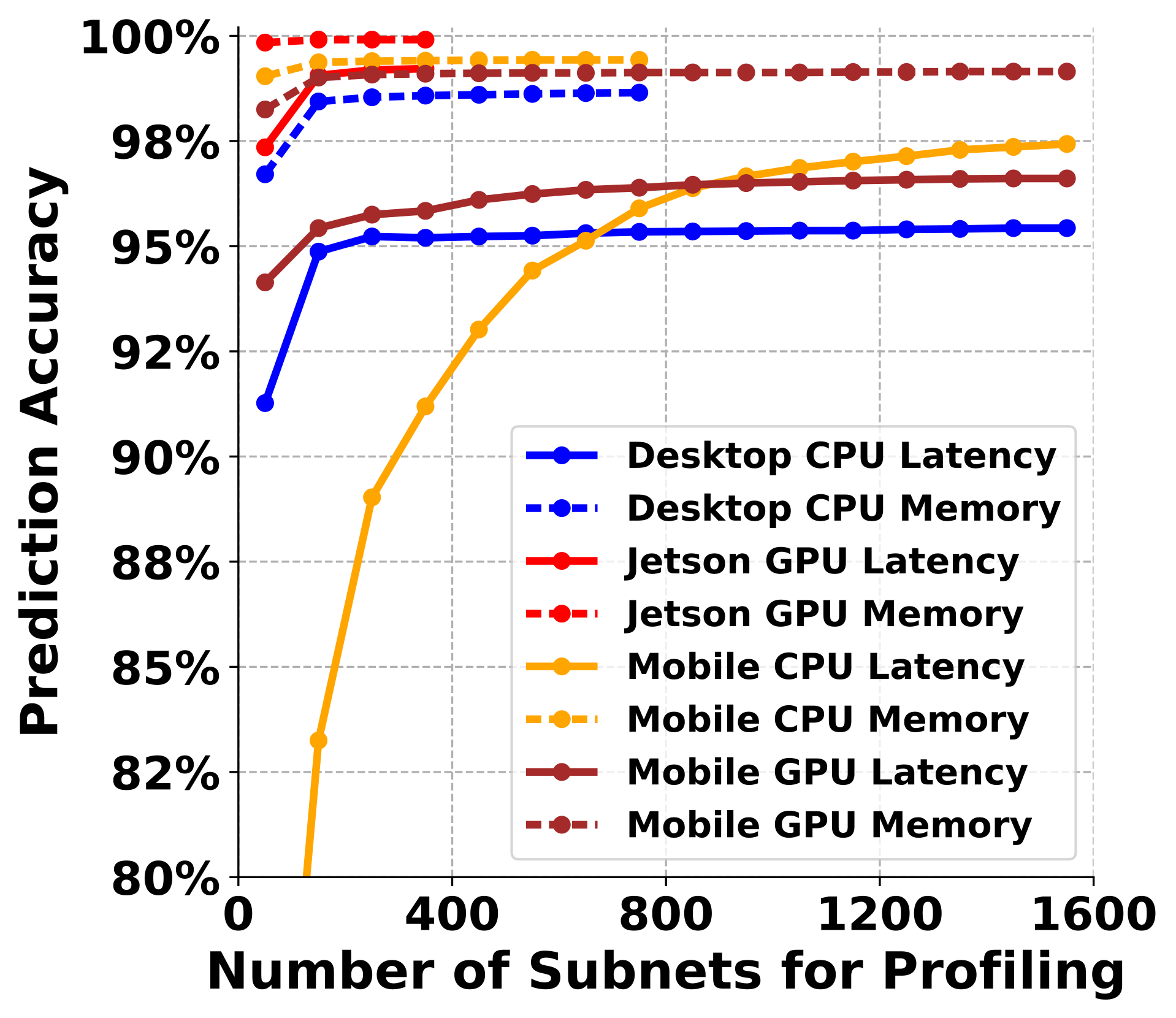}
      \vspace{-0.5cm}
      \caption{The accuracy of modeling for performance prediction by training with different \#subnets.}
      \label{fig:Acc vs Cost}
\end{minipage}
\end{figure}

\textbf{One-time Preparation Cost.}
The preparation cost of \name mainly came from the training of the modular supernet.
First, we analyzed the relation between the quality of generated model and number of training epochs. As depicted in Figure ~\ref{fig:EpochAcc}, \name requires more training epochs to gradually improve the quality of the generated models. This is associated with our training methodology. \name undergoes concurrent training on multiple tasks, with each task having access to only a fraction of the training data in each epoch. Consequently, it necessitates more training epochs to produce high-quality models for each individual task. Nonetheless, owing to the substantial inter-dependencies among tasks, their concurrent training has a synergistic effect, ensuring that \name's training overhead does not escalate significantly.





The performance predictor used for lightweight architecture search is also built during preparation. Figure~\ref{fig:Acc vs Cost} shows the performance modeling accuracy achieved with different number of subnets used for profiling and modeling (the supernet backbone is ResNet). 
As the number of subnets increases, the prediction accuracy shows an upward trend and reaches an asymptotic line. We can see that using 200$\sim$1000 subnets for profiling and training is sufficient to achieve good (higher than 95\%) latency and memory prediction accuracy.
Collecting the profiling data for one subnet took 70ms (Jetson GPU) to 985ms (Mobile CPU). Meanwhile, the time needed for fitting the performance model is less than few seconds. Therefore, the time to establish the performance prediction model in \name ranges from approximately 14s to 985s, which is negligible as an offline one-time process.





\textbf{On-device Serving Cost.}
The models generated by \name are normal static models, which do not produce any additional overhead at runtime.



\begin{figure}
    \centering
    \includegraphics[width=0.47\textwidth]{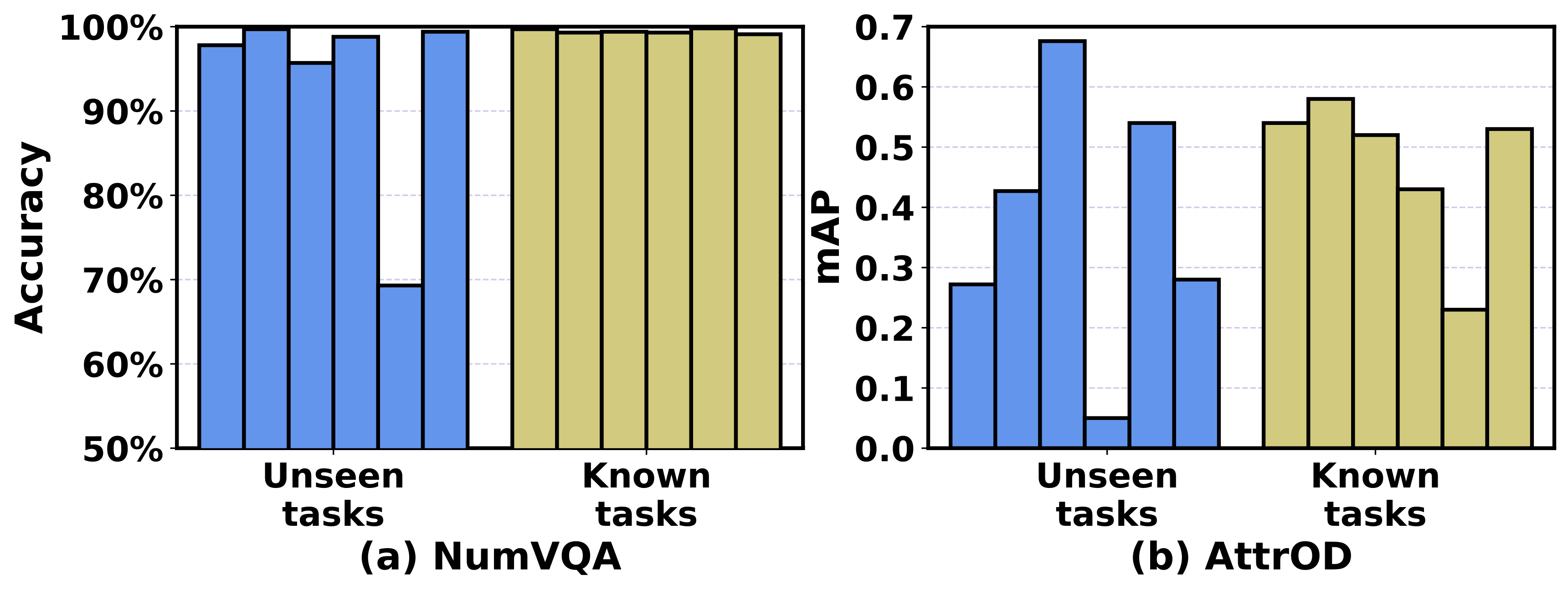}
    \vspace{-0.3cm}
    \caption{Accuracy or mAP for known \& unseen tasks in (a) NumVQA. (b) AttrOD.}
    \label{fig:UnseenTasks}
\end{figure}





\begin{figure}
    \centering
    \includegraphics[width=0.47\textwidth]{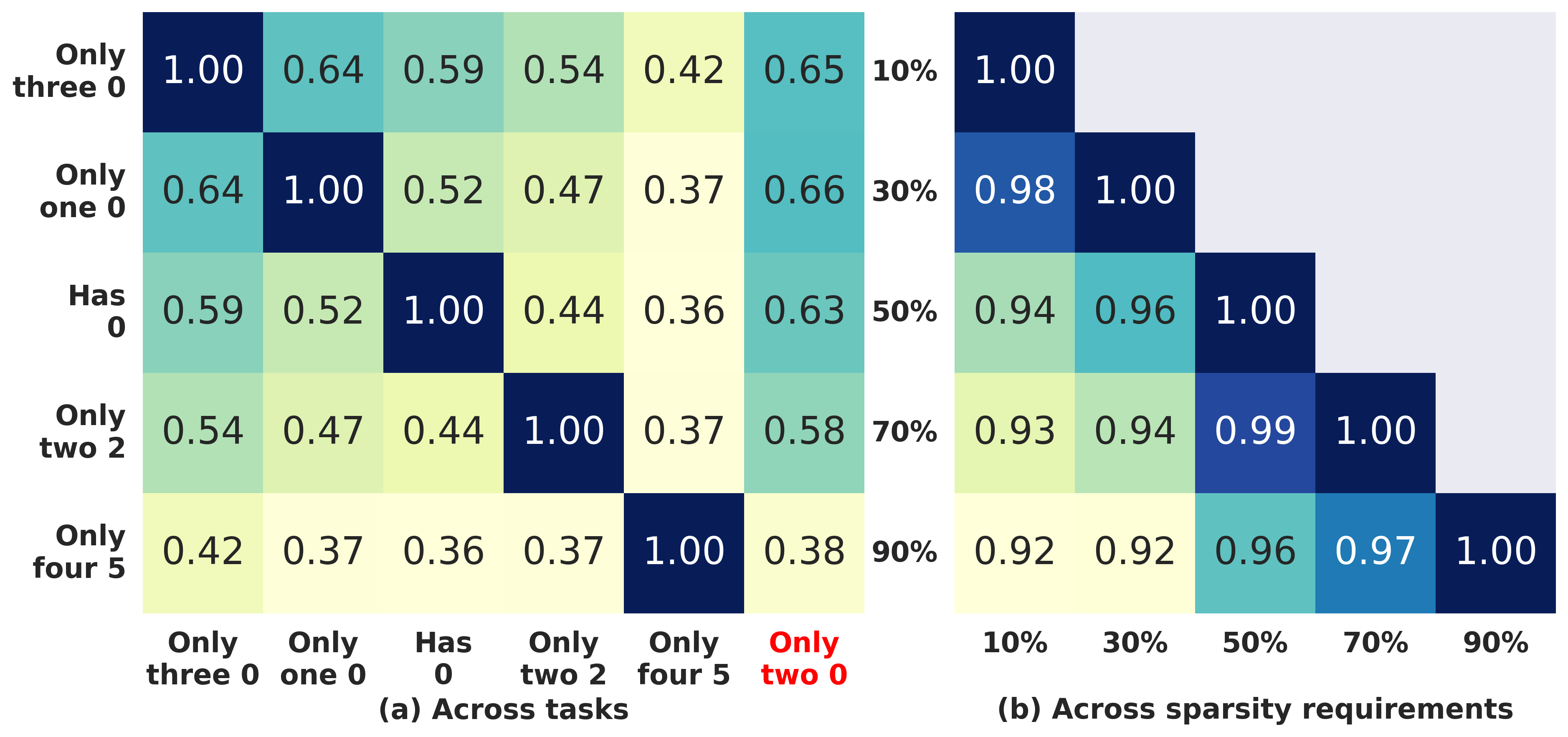}
    \vspace{-0.3cm}
    \caption{Gate similarity between different tasks (a) and between different activation limits (b) in NumVQA. The task marked in red is an unseen task.}
    \label{fig:UsageSimTwo}
\end{figure}


\subsection{Generalization to Unseen Tasks}
\label{eval:unseen-tasks}


In this section, we evaluate the generalization capability of \name to unseen tasks. Unseen tasks refer to the tasks that are not part of the training task set, yet they share the same combinatorial property space as the training tasks.

Figure \ref{fig:UnseenTasks} illustrates the accuracy of randomly selected unseen tasks in NumVQA and AttrOD. Although the accuracy for unseen tasks may exhibit a slight reduction as compared to that of known tasks and few unseen tasks may even yield a much lower accuracy, the majority of tasks demonstrate excellent accuracy, meaning that the modules in \name can be effectively assembled to handle new tasks without training data.
This result highlights the robust generalization capabilities of \name, as it can grasp the meanings of the properties comprising a task and comprehend the operations involved in their combinations.




We further analyzed the generalization capabilities of \name based on the similarity between the gate selections for different tasks. The result is presented in Figure~\ref{fig:UsageSimTwo}. The gate selections exhibit higher similarity when tasks are more alike (\eg `Only one 0' and `Only three 0'). Additionally, within a fixed task, increased proximity in the given activation limits leads to greater similarity in gate selections. This demonstrates the ability of \name to understand both the specified task and activation limit requirements, and map them to corresponding modules. This serves as the underpinning for its generalization abilities.

With this ability, \name can identify the most relevant modules for an unseen task and assemble a model with them. For example, the activated modules for the unseen `only two 0' task are similar with the modules of `only three 0' and `only one 0' in Figure~\ref{fig:UsageSimTwo}.
This leads to a high accuracy on the unseen task. Therefore, the generalization capability of \name is positively correlated with the number of tasks for training, which can lead to more useful modules and a more powerful assembler.






\

\section{Discussion}




Here we highlight some issues that warrant further discussion.

\textbf{Applicability to other types of tasks.} 
Currently, \name only support customizing models for tasks that are in a combinatorial space, which may limit its applicability to more generic usage scenarios.
To enable more flexible task spaces, we need to use a more powerful generative model as the assembler, which takes free-form task definition (\eg natural language task description) as the input, and operates on a larger set of neural modules.
Meanwhile, a large dataset containing the mappings between different tasks and corresponding input/output pairs is required to train the assembler and modules.
This is feasible according to the recent advances of pretrained large models (\eg ImageBind, ChatGPT, etc.), but training such a general-purpose model generator is very time-consuming and resource-intensive, which is impractical for most researchers.
We leave the development of such general-purpose \name for future work.


\textbf{Preparation cost of \name.}
The remarkable rapid model customization ability of \name comes at the cost of longer offline preparation time.
Specifically, training the modular supernet and the module assembler takes much longer time than training a normal static model.
This is because that \name needs to not only learn how to solve each individual task, but also how to decouple the modules and reassemble them to solve new tasks.
Considering the reduced marginal cost for supporting diverse edge scenarios, the one-time preparation cost is less significant.
Such benefits of marginal cost reduction are more valuable if \name supports more flexible task spaces.

\section{Conclusion}

This paper proposes a novel approach for rapid customization of deep learning models for diverse edge scenarios. With a holistic design with a modular supernet, a module assembler, and a lightweight architecture searcher, we are able to achieve rapid model customization for deverse edge tasks and resource constraints. Experiments have demonstrated excellent model generation quality and speed of our approach. We believe our work has enabled a new and important generative model customization experience.



\balance
\bibliographystyle{ACM-Reference-Format}
\bibliography{reference}

\end{document}